\documentclass{article}

\usepackage{microtype}
\usepackage{graphicx}
\usepackage{subcaption}
\usepackage{booktabs} % for professional tables

\usepackage{hyperref}

\usepackage[preprint]{icml2026}

\usepackage{amsmath}
\usepackage{amssymb}
\usepackage{mathtools}
\usepackage{amsthm}
\usepackage{enumitem}
\usepackage[capitalize,noabbrev]{cleveref}

\newcommand{\appdotleader}{\leaders\hbox to 0.55em{\hss.\hss}\hfill}
\newcommand{\appcontentsline}[2]{%
  \par\addvspace{0.45em}%
  \noindent
  \makebox[2.8em][l]{\hyperref[#1]{\textbf{\ref*{#1}}}}%
  \hyperref[#1]{\textbf{#2}}%
  \nobreak\appdotleader
  \nobreak\makebox[2.4em][r]{\hyperref[#1]{\textbf{\pageref*{#1}}}}\par
}
\newcommand{\appsubcontentsline}[2]{%
  \par\addvspace{0.1em}%
  \noindent\hspace*{1.7em}%
  \makebox[3.1em][l]{\hyperref[#1]{\ref*{#1}}}%
  \hyperref[#1]{#2}%
  \nobreak\appdotleader
  \nobreak\makebox[2.4em][r]{\hyperref[#1]{\pageref*{#1}}}\par
}
\theoremstyle{plain}

\theoremstyle{definition}

\theoremstyle{remark}

\usepackage[textsize=tiny]{todonotes}

\usepackage[dvipsnames]{xcolor}
\usepackage{colortbl}
\usepackage[most,skins,theorems]{tcolorbox}
\tcbuselibrary{theorems, breakable, skins}
\newtcolorbox{highlight-block}[1][]{
    colback=gray!5,
    colframe=gray!60!black,
    boxrule=0.5pt,
    breakable,
    left=6pt,
    right=6pt,
    top=6pt,
    bottom=6pt,
    arc=2pt,
    #1
}
\newtcolorbox{theorem-block}[1][]{
    colback=blue!3!white,
    colframe=blue!60!black,
    boxrule=0.5pt,
    breakable,
    left=6pt,
    right=6pt,
    top=6pt,
    bottom=6pt,
    arc=2pt,
    #1
}
\newtcolorbox{proposition-block}[1][]{
    colback=blue!3!white,
    colframe=blue!60!black,
    boxrule=0.5pt,
    breakable,
    left=6pt,
    right=6pt,
    top=6pt,
    bottom=6pt,
    arc=2pt,
    #1
}
\newtcolorbox{discussion-block}[1][]{
    colback=orange!4!white,  % 极浅的背景色，几乎看不出
    colframe=white,
    boxrule=0.0pt,           % 超细线框（原0.5pt）
    breakable,
    left=4pt,                % 更小的内边距
    right=4pt,
    top=4pt,
    bottom=4pt,
    arc=1pt,                 % 更小圆角
    boxsep=0pt,              % 紧凑排版
    #1
}

\tcbset{
  aibox/.style={
    width=\linewidth,
    top=7pt,
    bottom=2pt,
    colback=blue!6!white,
    colframe=black,
    colbacktitle=black,
    enhanced,
    center,
    attach boxed title to top left={yshift=-0.1in,xshift=0.15in},
    boxed title style={boxrule=0pt,colframe=white,},
  }
}
\newtcolorbox{AIbox}[2][]{aibox,title=#2,#1}

\newcommand{\highlight}[1]{{\color{red!20!violet}#1}}

\newcommand{\coloredhref}[3][blue]{%
  \href{#2}{\textcolor{#1}{#3}}%
}

\definecolor{table-blue}{RGB}{173, 216, 230}
\definecolor{darkgreen}{rgb}{0.0, 0.5, 0.0}
\definecolor{darkblue}{rgb}{0, 0, 0.5}
\definecolor{paleviolet}{HTML}{E1EEFC}
\definecolor{burgundy}{rgb}{0.5, 0.0, 0.13}

\icmltitlerunning{Depth-Breadth Synergy in RLVR: Unlocking LLM Reasoning Gains with Adaptive Exploration}

\begin{document}

% \author{
% ~~~~~~Zhicheng Yang\textsuperscript{1}~~
% Zhijiang Guo\textsuperscript{1,2}~~
% Yinya Huang\textsuperscript{3}~~
% Yongxin Wang\textsuperscript{6}~~ 
% Dongchun Xie\textsuperscript{5}
% \\
% ~~~~~~~~~~~~~~~~~~~~~~~~~~~~~~~~~~~~~~~~~\textbf{Yiwei Wang\textsuperscript{4}~~Xiaodan Liang\textsuperscript{5,6}~~Jing Tang\textsuperscript{1,2}}\thanks{Corresponding author: Jing Tang.} \\
% ~~~~~~~~~~~~~~~~~~~~~$^1$The Hong Kong University of Science and Technology (Guangzhou) \\
% ~~~~~~~$^2$The Hong Kong University of Science and Technology ~~
% $^3$ETH AI Center, ETH Zurich ~~ \\
% ~~~~~~~~~~~~~~~~~~$^4$University of California, Merced ~~
% $^5$Sun Yat-sen University ~~
% $^6$MBZUAI ~~\\
% ~~~~~~~~~~~~~~~~~~~~~~~~~~~~~~~~~~~~~~~~~~~~~~~~~~~~~~~~\texttt{yangzhch6@gmail.com} \\ \\
% ~~~~~~~~~~~~~~~~~~~~~~~~~~~\textit{Project Repo}: \href{https://github.com/yangzhch6/DARS}{ \ttfamily https://github.com/yangzhch6/DARS}
% % ~~~~~~~~\texttt{\{yangzhch6, cartusguo, xdliang328, wangyw.evan\}@gmail.com} \\ 
% % ~~~~~~~~~~~~~~~~~~~~~~~~~~~~\texttt{yinya.huang@hotmail.com, jingtang@ust.hk} 
% }

\onecolumn
{
\vspace{-5mm}
  \icmltitle{Depth-Breadth Synergy in RLVR: \\
  Unlocking LLM Reasoning Gains with Adaptive Exploration}




  \vspace{-6mm}
  \begin{icmlauthorlist}
    \icmlauthor{Zhicheng Yang}{hkustgz}
    \icmlauthor{Zhijiang Guo}{hkustgz,hkust}
    \icmlauthor{Yinya Huang}{eth}
    \icmlauthor{Yongxin Wang}{MBZUAI}
    \icmlauthor{Dongchun Xie}{sysu}
    \icmlauthor{Hanhui Li}{sysu}
    \icmlauthor{Yiwei Wang}{ucm}
    \icmlauthor{Xiaodan Liang}{MBZUAI,sysu}
    \icmlauthor{Jing Tang}{hkustgz,hkust}
  \end{icmlauthorlist}

  \icmlaffiliation{hkustgz}{HKUST-GZ}
  \icmlaffiliation{hkust}{HKUST}
  \icmlaffiliation{sysu}{Sun Yat-Sen University}
  \icmlaffiliation{eth}{ETH AI Center, ETH Zurich}
  \icmlaffiliation{MBZUAI}{MBZUAI}
  \icmlaffiliation{ucm}{University of California, Merced}

  % \icmlcorrespondingauthor{}{yangzhch6@gmail.com}

  \icmlkeywords{Machine Learning, ICML}
  \vspace{-4mm}
  \center{\texttt{yangzhch6@gmail}}
  \vspace{-3mm}
  % \center{\textit{Project Repo}: \href{https://github.com/yangzhch6/DARS}{\ttfamily https://github.com/yangzhch6/DARS}}
  \center{\textit{\textbf{Project Repo}}: \coloredhref[burgundy]{https://github.com/yangzhch6/DARS}{\ttfamily \textbf{https://github.com/yangzhch6/DARS}}}
  \vspace{-2mm}
  \vskip 0.2in
}



% Use ONE of the following lines. DO NOT remove the command'.

\printAffiliationsAndNotice{}  % no special notice (required even if empty)

\begin{abstract}
    \vspace{-1mm}
  Reinforcement Learning with Verifiable Reward (RLVR) is a powerful method for enhancing the reasoning abilities of Large Language Models, but its full potential is limited by a lack of exploration in two key areas: \textbf{Depth} (the difficulty of problems) and \textbf{Breadth} (the number of training instances). Our analysis of the popular GRPO algorithm reveals a bias that down-weights difficult, low-accuracy problems, which are crucial for improving reasoning skills. To address this, we introduce \textbf{D}ifficulty \textbf{A}daptive \textbf{R}ollout \textbf{S}ampling (\textbf{DARS}), a method that re-weights difficult problems by using targeted, multi-stage rollouts. DARS increases the number of rollout outcomes for these harder problems according to our proposed re-balancing schedules and leads to consistent gains in \textit{Pass@K}. We discovered that increasing rollout size alone does not improve performance and may actually impair it. In contrast, scaling the batch size to increase breadth via full-batch updates significantly boosted Pass@1 metrics. This improvement stems from higher token-level entropy, ensuring robust exploration and minimized gradient noise.
  We further present DARS-Breadth, a combined approach that uses DARS with a large breadth of training data. This method demonstrates simultaneous gains in both \textit{Pass@K} and \textit{Pass@1}, confirming that depth (adaptive exploration) and breadth (scaling iteration instances) are orthogonal and complementary dimensions for unlocking the full power of RLVR.

\vspace{-2mm}

\begin{figure}[htbp] %     % width=0.99\textwidth
    \centering
    \includegraphics[width=0.88\linewidth]{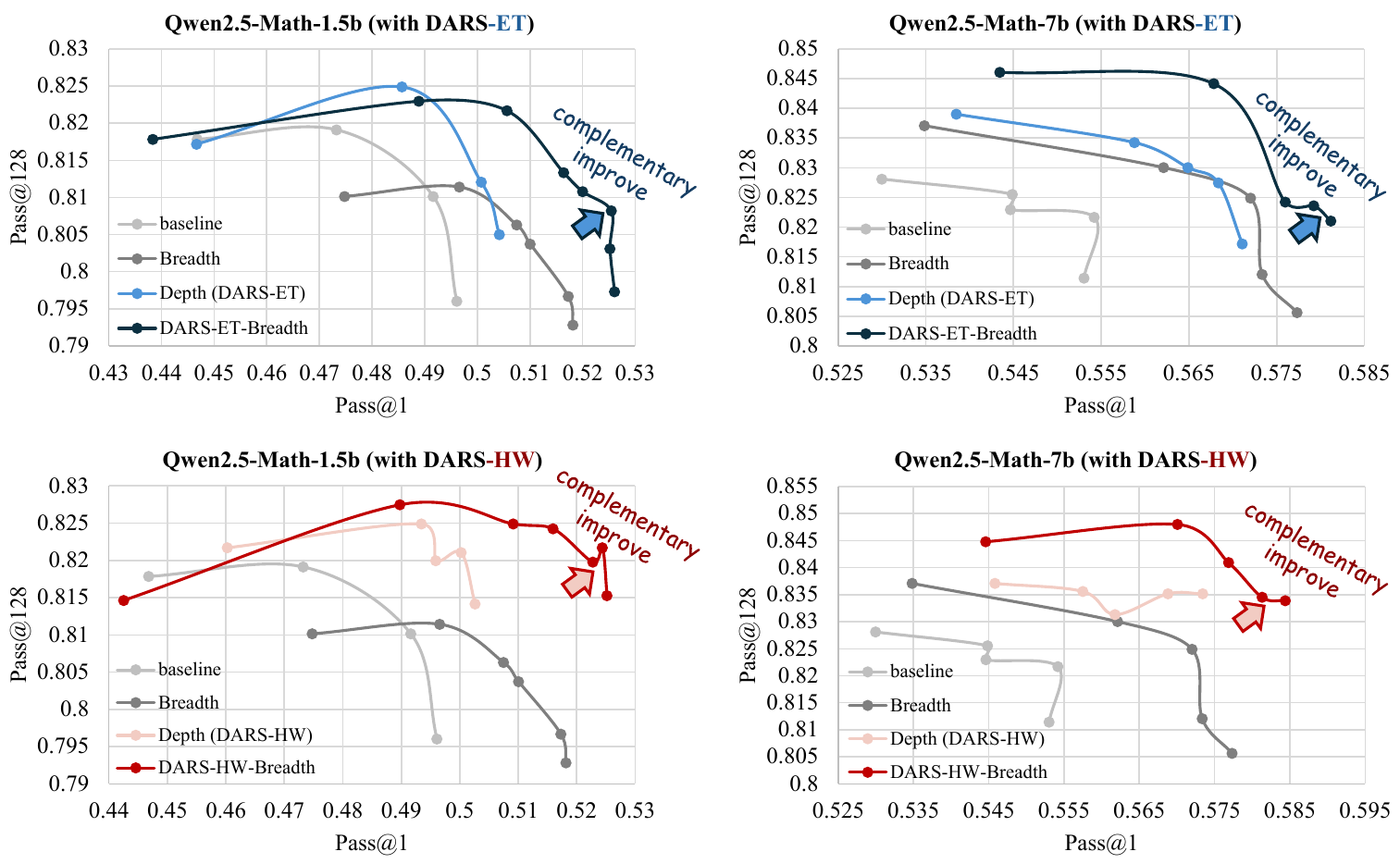}
    \vspace{-3mm}
    \caption{Depth and Breadth Synergy for \textit{Pass@1} and \textit{Pass@K} (K=128) performance.}\label{fig:D_B_128}
\end{figure}

\end{abstract}

\twocolumn[]

% \vspace{-7mm}
\section{Introduction}

% \vspace{-2mm}
The emergence of reasoning-centric Large Language Models (LLMs) exemplified by OpenAI-o1~\citep{jaech2024openaio1}, DeepSeek-R1~\citep{guo2025deepseek-r1}, and Kimi-1.5~\citep{team2025kimi}, has pushed the frontier of LLM capability, especially for demanding tasks in complex reasoning such as mathematics and programming. 
Unlike conventional instruction tuning that relies on human-labeled data or RLHF pipelines that demand an auxiliary, well-trained reward model~\citep{ouyang2022traininglanguagemodelsfollow,achiam2023gpt4,grattafiori2024llama3}, this leap is driven by large-scale Reinforcement Learning with Verifiable Rewards (RLVR;~\citealt{guo2025deepseek-r1, zeng2025simplerl}) for which correctness can be automatically and deterministically checked. The rewards of RLVR are granted solely when a model’s output matches the ground-truth mathematical answer or passes all unit tests for code, allowing scalable verification without manual labeling. 
% This framework has drawn widespread attention for its elegant simplicity and proven effectiveness. Some studies~\citep{guo2025deepseek-r1, zeng2025simplerl} have found that RLVR enables LLMs to autonomously develop new reasoning patterns such as self-reflection and can even elicit slow-thinking capabilities directly from the base model. 
RLVR is now regarded as a promising path toward self-evolving LLMs, potentially bringing us closer to more powerful intelligence. 

However, existing RLVR frameworks inadequately address the interplay between exploration depth (difficulty scaling) and breadth (iteration instance quantity scaling), which leads to insufficient performance gain for both \textit{Pass@1} and \textit{Pass@K}. In this paper, we conduct a systematic analysis of these two under-exploited dimensions in RLVR.

\begin{figure*}[htbp] % htbp    % width=0.99\textwidth
    \centering
    \includegraphics[width=0.9\linewidth]{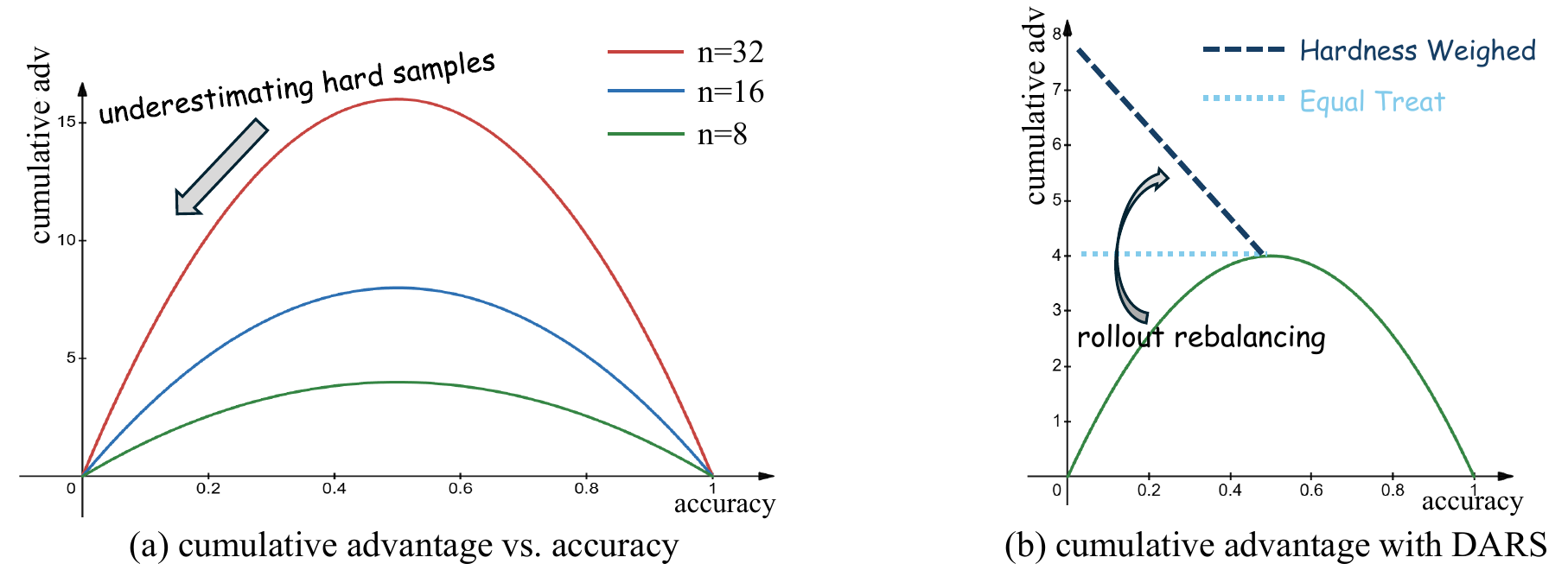}
    \caption{Statistical results of cumulative advantage. Group relative advantage calculation methods underestimate high-difficulty problems. $n$ denotes group size.}\label{fig:abstcum_advract}
    % \vspace{-3mm}
\end{figure*}

% depth
% For the dimension of \textbf{depth}, our investigation reveals that existing methods of GRPO~\citep{deepseekmathgrpo} and its variants~\citep{dapo,liu2025understanding}, while adept at estimating the advantage of a single rollout, are undermined by distorted cumulative advantage at the group-level that allocates disproportionate attention to instances of medium-difficulty while neglecting the high-difficulty instances indispensable for complex reasoning, as illustrated in Figure \ref{fig:abstcum_advract}.
For the dimension of \textbf{depth}, our investigation reveals that existing methods of GRPO~\citep{deepseekmathgrpo} and its variants~\citep{dapo,liu2025understanding}, while adept at estimating the advantage of a single rollout, are undermined by a distorted cumulative advantage at the group level. This distortion disproportionately allocates attention to instances of medium difficulty, neglecting high-difficulty instances indispensable for complex reasoning, as illustrated in Figure \ref{fig:abstcum_advract}.
% For the dimension of \textbf{depth}, our investigation reveals that existing methods of GRPO~\citep{deepseekmathgrpo} and its variants, while adept at exploiting verifiable reward signals, are undermined by a skewed advantage-weighting scheme that lavishes attention on medium-accuracy instances while sidelining the low-accuracy, high-difficulty instances indispensable for complex reasoning. 
This bias fundamentally limits depth, the hardest problems a model can learn to solve, and constrains \textit{Pass@K} performance. 
To counteract this depth neglect, we propose Difficulty-Adaptive Rollout Sampling (\textbf{\textsc{DARS}}). DARS performs a lightweight first-stage rollout to estimate per-problem accuracies, then allocates additional compute via targeted multi-stage rollouts to low-accuracy problems. By expanding sampling on hard problems, DARS re-weights the cumulative advantage, making it easier for LLMs to learn `deep' samples and improving \textit{Pass@K} performance.

We further identify \textbf{breadth} as the instance quantity consumed in a single iteration. We observe that breadth has a significant impact on the LLM's performance and continuous exploration capability, as shown in Figure \ref{fig:breadth-entropy}. 
In DARS, the dynamic rollout allocation \(\Delta n_j\) yields ragged per-question rollout counts that break standard PPO minibatch chunking, so we adopt full-batch updates for multiple PPO epochs as a compatible training design. 
Based on this, we significantly increase the training batch size. This seemingly simple change dramatically improves \textit{Pass@1} and sustains high token-level entropy throughout training, suggesting that breadth acts as implicit entropy regularization that delays premature convergence. Importantly, the gains from breadth are complementary to those from depth: we present \textbf{DARS-Breadth} that combines our DARS with large-breadth training, producing simultaneous boosts in both \textit{Pass@K} and \textit{Pass@1}. Our contributions can be summarized as follows:
% \vspace{-1mm}
\begin{itemize}[leftmargin=*]
    % \vspace{-2mm}
    \item \textbf{Cumulative Advantage Bias}: We conduct a systematic analysis on depth and breadth in RLVR, and uncover the depth bias in GRPO: cumulative advantage underweights low-accuracy, high-difficulty samples, capping \textit{Pass@K} performance.

    % \vspace{-2mm}
    \item \textbf{Difficulty Adaptive Rollout Sampling}: We introduce DARS, which reallocates compute to the hard problems via multi-stage rollout sampling. DARS re-weights the cumulative advantage distribution and quantitatively expands the sparse reward signals for difficult problems. We further show that the ET and HW schedules recover \textbf{Log-Odds} and \textbf{Maximum Likelihood} objectives, respectively. In practice, both schedules significantly improve \textit{Pass@K} performance. 
    % We elaborate on the impact of cumulative advantage on the gradient norm in Appendix \ref{app:gradient_norm}.
    
    % \vspace{-2mm}
    \item \textbf{Depth-Breadth Synergy}: We then illustrate that large breadth matters for the \textit{Pass@1} performance in RLVR.
    By combining DARS with large breadth scaling, we further reveal the complementarity of Depth and Breadth and acquire simultaneous boosts in both \textit{Pass@K} and \textit{Pass@1} performance.

\end{itemize}

% \vspace{-3mm}

\section{Understanding RLVR from Depth and Breadth}
\label{sec:analysis}

% \vspace{-3mm}
\begin{figure*}[t] % htbp    % width=0.99\textwidth
    \centering
    \includegraphics[width=1\linewidth]{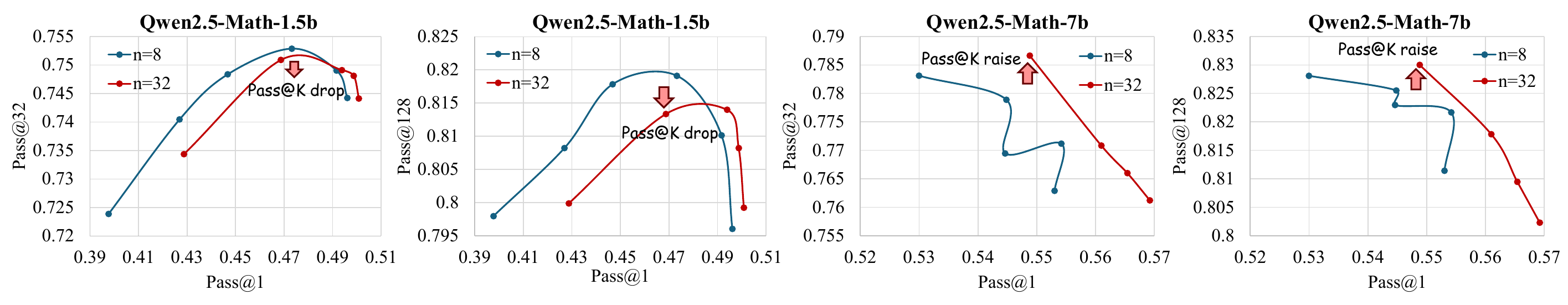}
    % \vspace{-6mm}
    \caption{Training dynamics of \textit{Pass@1} and \textit{Pass@K} performance of Qwen2.5-Math-1.5B and -7B with different rollout size.}\label{fig:depth-naive}
\end{figure*}
% \vspace{-10mm}

\subsection{Depth: The Hardest Problem Sampled in RLVR}
We first identify \textbf{Depth} as the hardest problem that can be correctly answered in the RLVR training process. In the GRPO training process, groups whose entire rollouts yield incorrect answers suffer from gradient vanishing. Hence, sampling high-difficulty questions with correct reasoning paths is crucial for LLM training. We first show that merely increasing rollout size does not consistently yield significant gains in \textit{Pass@K} performance, and sometimes can even be harmful. We then quantify GRPO’s cumulative advantage and highlight its under-weighting of high-difficulty samples.

\noindent
\textbf{Naive Scaling of Rollout Size Benefits \textit{Pass@1}, But Not Necessarily \textit{Pass@K}.}
We present the training dynamics of \textit{Pass@1} and \textit{Pass@K} performance during the RLVR training process in Figure \ref{fig:depth-naive}. Enlarging the rollout size allows the sampling of correct solutions to hard problems during training. We originally assumed this would benefit \textit{Pass@K} performance; however, experimental results show that this is not always the case. We find that Qwen2.5-Math-7B can significantly benefit from an increased rollout size, whereas for Qwen2.5-Math-1.5B, naively scaling rollout size can even harm \textit{Pass@K} performance.

\noindent
\textbf{Cumulative Advantage Bias in GRPO Variants hinders the improvement of \textit{Pass@K}.}
In the GRPO framework, the advantage estimation is derived by normalizing binary rewards:
\begin{equation}
\hat{A}_{i}^{std} = \frac{r_i - u}{\sigma},~~~~~\hat{A}_{i}^{nostd} = r_i - u, 
\label{eq:grpo_adv}
\end{equation}
where $r_i$ is the binary reward of $i_{th}$ rollout, $u$ is the mean value of the group rewards $u=\text{mean}(\{R_i\}_{i=1}^G)$ and $\sigma$ is the standard deviation of the group rewards $\sigma = \text{std}(\{R_i\}_{i=1}^G)$. In the case of binary rewards, $u$ also represents the accuracy of LLM rollouts. Dr.~GRPO~ \citep{liu2025understanding} removes the standard-deviation term from the advantage computation to eliminate question-level difficulty bias, and demonstrates its superiority through extensive experiments. Consequently, the experiments reported in this study were conducted primarily though the Dr.~GRPO methodology. We show more results of std-based advantage in Appendix \ref{apd:std}.

For a group $G$ with rollout size $N$, we define the \textbf{cumulative advantage} of a group as the sum of the absolute values of sample advantages: $\mathcal{A}_{\text{group}} \;=\; \sum_{i=1}^{G}\,|\hat{A}_{i}|$. The cumulative advantage reflects how much the algorithm weights each sample. Specifically, for Dr.~GRPO, 
% \begin{equation}
% \mathcal{A}_{\text{group}}^{std} = 2 N \sqrt{u(1-u)}, ~~~\mathcal{A}_{\text{group}}^{nostd} = 2 N u (1-u), 
% \label{eq:cum_adv}
% \end{equation}
\begin{equation}
\mathcal{A}_{\text{group}} = 2 N u (1-u).
\label{eq:cum_adv}
\end{equation}
The cumulative advantage functional curve is plotted in Figure~\ref{fig:abstcum_advract}. As shown in the figure, group-based advantage computation funnels its weight toward problems of medium difficulty while largely overlooking those that are highly difficult. This bias limits the \textit{Pass@K} performance of RLVR. 
% To address this issue, we propose Difficulty Adaptive Rollout Sampling (DARS) to balance the cumulative advantage of hard samples, as detailed in Section \ref{sec:dars}.

\begin{figure}[t] % htbp    % width=0.99\textwidth
    \centering
    \includegraphics[width=1.0\linewidth]{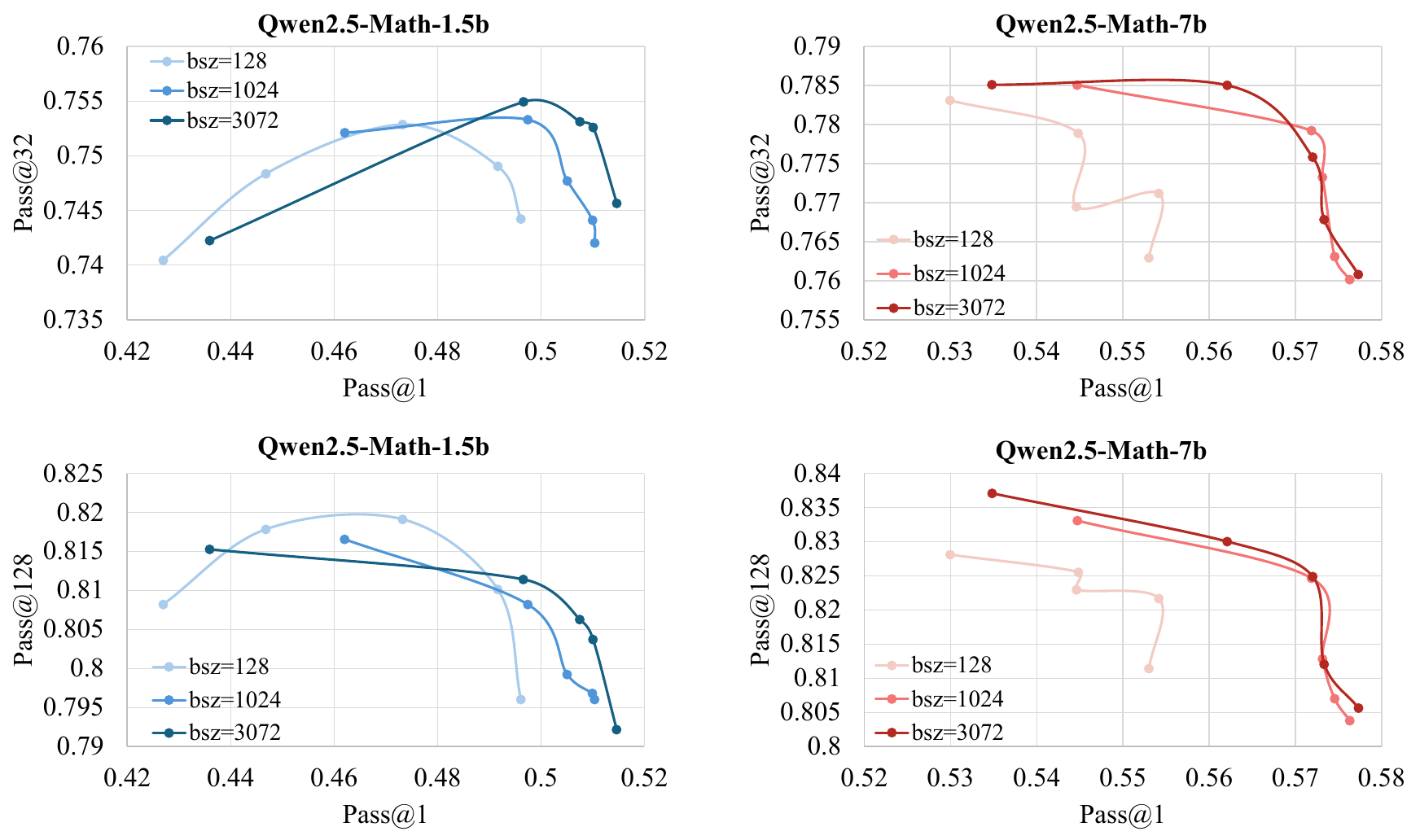}
    \caption{Training dynamics of \textit{Pass@1} and \textit{Pass@K} performance of Qwen2.5-Math-1.5B and -7B with different batch size.}\label{fig:breadth-naive}
\end{figure}

\subsection{Breadth: Iteration Instance Quantity in RLVR}
\label{sec:analysis-breadth}
% We identify \textbf{Breadth} as the quantity of instances used in each iteration of the RLVR process. In this subsection, we discuss the \textit{Pass@1} performance benefits of simply increasing the batch size for the RLVR process. 

We define \textbf{Breadth} as the number of instances used per iteration of the RLVR process. We'll show how increasing the batch size for the RLVR process improves the \textit{Pass@1} performance.

% In section \ref{sec:breadth-method}, we further discussed the impact of replacing the PPO mini batch with the full batch update for multiple PPO epochs.

\noindent
\textbf{Breadth Matters for \textit{Pass@1} Performance.}
Most studies~\citep{liu2025understanding,liu2025there,luffy,fu2025srft} conventionally set the batch size to 128. In this subsection, we drastically increase the batch size to 3072 and plot the training dynamics of \textit{Pass@1} and \textit{Pass@32} performance in Figure \ref{fig:breadth-naive}. Although large-batch optimization is not new in the broader deep learning literature, its effect in RLVR is notable: naively increasing the batch size brings a \textit{Pass@1} improvement for all models, yet it harms the \textit{Pass@128} performance of Qwen2.5-Math-1.5B.
We consider that increasing the quantity of instances used in each iteration makes the gradient direction more accurate and reduces the impact of noise, thereby improving \textit{Pass@1} performance.

\begin{figure}[htbp] % htbp    % width=0.99\textwidth
    \centering
    \includegraphics[width=0.99\linewidth]{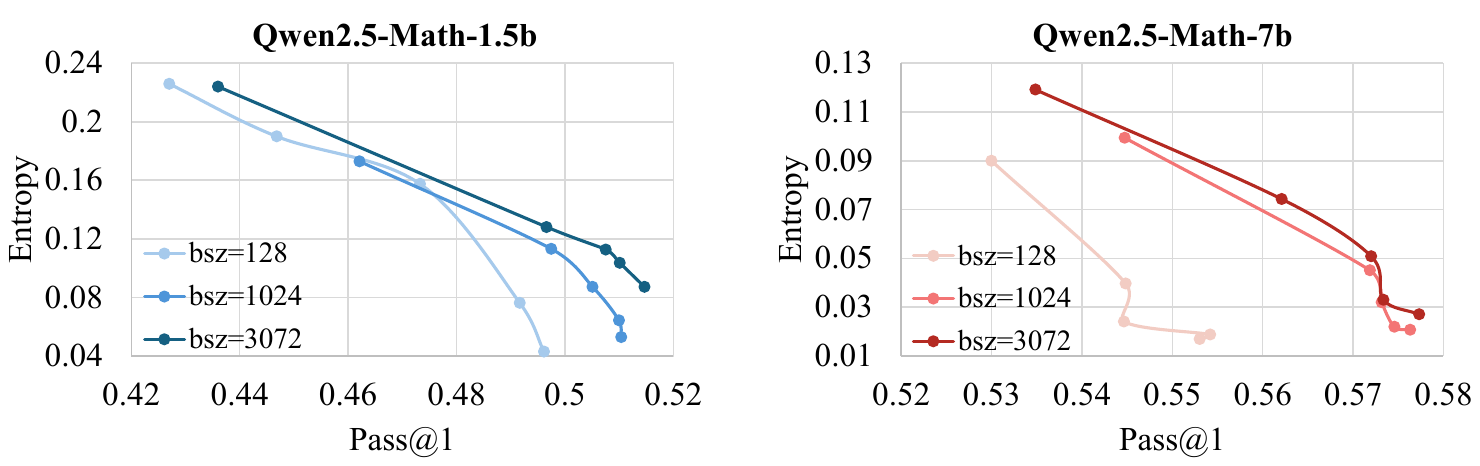}
    \caption{Training dynamics of \textit{Pass@1} performance and token entropy for Qwen2.5-Math-1.5B and Qwen2.5-Math-7B.}\label{fig:breadth-entropy}
\end{figure}

\noindent
\textbf{Breadth Sustains Entropy for Model Exploration.}

High token entropy in LLMs indicates strong exploration capabilities. Our analysis shows a relationship between \textit{Pass@1} and token entropy during training. As illustrated in Figure \ref{fig:breadth-entropy}, increased training breadth enables LLMs to achieve higher entropy at a given Pass@1 accuracy. We consider a large training breadth acts as a form of entropy regularization, preventing premature convergence and boosting \textit{Pass@1} performance while maintaining high entropy.

\begin{figure*}[t] % htbp    % width=0.99\textwidth
    \centering
    \includegraphics[width=1.0\linewidth]{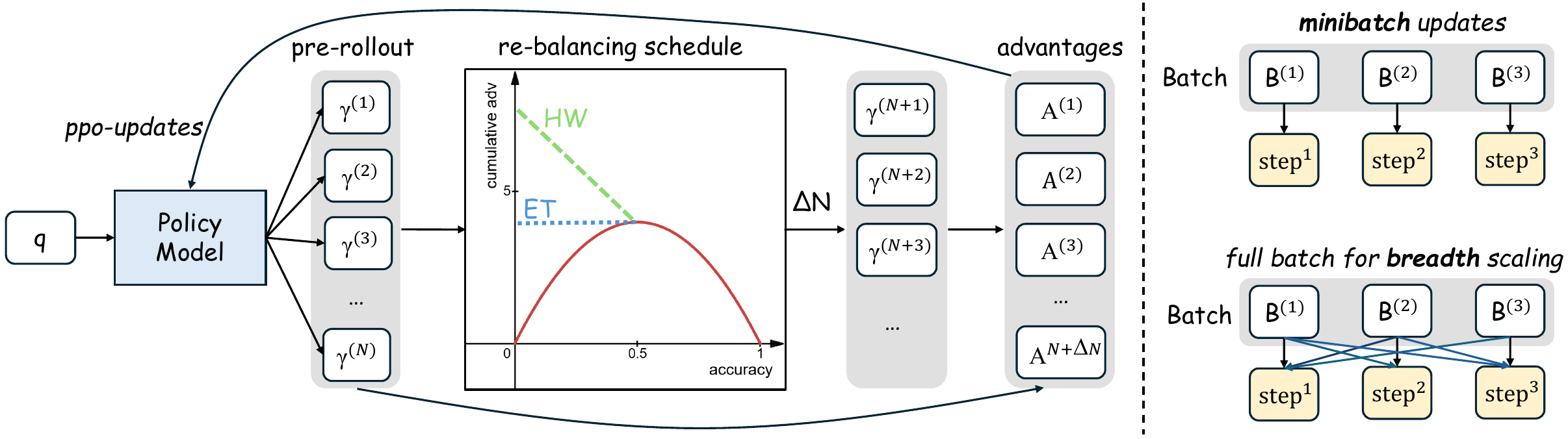}
    \caption{The overall training framework of our Difficulty Adaptive Rollout Sampling (\textbf{DARS}) with breadth scaling. Our DARS consists of 2 phases: 1) a pre-rollout stage to evaluate the difficulty of the given question, and 2) a re-balancing rollout stage to adjust the cumulative advantage. For breadth scaling, we replace the PPO minibatch with the full batch with multiple PPO epochs.}\label{fig:framework}
\end{figure*}

\section{Methodology}
In Section \ref{sec:analysis}, we analyze the bias inherent in group-based advantage computation. To solve this issue, we introduce Difficulty Adaptive Rollout Sampling (\textbf{DARS}), which rebalances the cumulative advantage via multi-stage sampling. By combining this depth-oriented mechanism with a compatibility-driven large-breadth training scheme, we further obtain \textbf{DARS-Breadth}, which improves both \textit{Pass@1} and \textit{Pass@K}.

\subsection{Difficulty Adaptive Rollout Sampling (DARS)}
\label{sec:dars}

As shown in Figure~\ref{fig:framework}, given a data batch \(\mathcal{B}=\{q_j\}_{j=1}^{M}\) of reasoning questions, DARS operates in two phases: (i) \textbf{pre-rollout difficulty estimation} that assigns to each question \(q_j\) a scalar difficulty score \(x_j\in[0,1]\); and (ii) \textbf{multi-stage rollout re-balancing} that dynamically decides how many additional trajectories \(\Delta n_j\) shall be allocated to \(q_j\) so that the cumulative advantage for low-accuracy problems is up-weighted. To simplify the subsequent formula representation, we define 
\begin{equation}
\mathcal{S}(\hat{a}_j) = 2\hat{a}_j(1-\hat{a}_j).
\label{eq:s_std}
\end{equation}

\noindent
\textbf{Phase 1: Pre-Rollout Difficulty Estimation.}  
For every \(q_j\), we draw a light first-stage rollout consisting of \(N^{pre}\) independent trajectories \(\{\tau_{j}^{(i)}\}\).  
Let the per-trajectory reward be binary, \(r_{j}^{(i)}\in\{0,1\}\).  
We define the empirical accuracy 
\begin{equation}
\hat{a}_j=\frac{1}{N^{pre}}\sum_{i=1}^{N^{pre}}r_{j}^{(i)}.
\label{eq:acc}
\end{equation}
The difficulty score is then set to the complementary accuracy $x_j=1-\hat{a}_j$, so that \(x_j\approx 1\) for the hardest problems and \(x_j\approx 0\) for the easiest ones. 

\noindent
\textbf{Phase 2: Multi-Stage Rollout Re-Balancing.}  
Let \(\mathcal{A}_{\text{group}}^N(u)\) denote the cumulative advantage under GRPO for a group whose average accuracy is \(u\) with rollout size $N$.  
We aim to reallocate \(\Delta N\) additional trajectories across the mini-batch so that the \textbf{effective} cumulative advantage for each question becomes an increasing function of its difficulty.  
To control the computing cost, we cap the rollout sampling upper limit at $N^{\mathrm{max}}$.
To this end, we design two rebalancing schedules.

\textbf{Schedule 1: Equal-Treatment (ET).}  
For every question \(q_j\) we enforce the rebalanced cumulative advantage as:
\begin{equation}
\mathcal{A}_{\text{group}}^{ET}(q_j)=\mathcal{A}_{\text{group}}^{N^{pre}}(0.5).
\label{eq:A_ET}
\end{equation}
We raise the cumulative advantage of all difficulty problems ($\hat{a}_j < 0.5$) to the level achieved by a medium-difficulty problem with accuracy \(\hat{a}_j=0.5\).  
The required extra trajectories are
\begin{equation}
\Delta n_j^{\text{ET}}=\mathrm{min}(
\left\lceil
\frac{\mathcal{A}_{\text{group}}^{N^{pre}}(0.5)-\mathcal{A}_{\text{group}}^{N^{pre}}(\hat{a}_j)}
{\mathcal{S}(\hat{a}_j)}
\right\rceil, N^{\mathrm{max}} - N^{pre}).
\label{eq:N_ET}
\end{equation}

\textbf{Schedule 2: Hardness-Weighted (HW).}  
We now impose a monotonically increasing re-weighting that allocates more rollouts to lower-accuracy problems:  

\begin{equation}
\mathcal{A}_{\text{group}}^{HW}(q_j)= 2x_j \mathcal{A}_{\text{group}}^{N^{pre}}(0.5).
\label{eq:A_HW}
\end{equation}

This yields

\begin{equation}
\begin{aligned}
\Delta n_j^{\text{HW}}=&\mathrm{min}(
\left\lceil
\frac{2x_j \cdot \mathcal{A}_{\text{group}}^{N^{pre}}(0.5) - \mathcal{A}_{\text{group}}^{N^{pre}}(\hat{a}_j)}
{\mathcal{S}(\hat{a}_j)}\right\rceil, 
\\& ~~~~~~~~~~~~~~~~~~~~~~~~~~~~~~~~~~~~~~~~~N^{\mathrm{max}} - N^{pre}).
\end{aligned}
\label{eq:N_HW}
\end{equation}

\textbf{Implicit Prompt-Level Objectives of ET/HW Schedules.}
Beyond rebalancing cumulative advantage, the ET/HW schedules also induce distinct \emph{prompt-level} optimization targets. Up to constant scaling, Dr. GRPO optimizes \textbf{Pass@1}, DARS-ET optimizes the \textbf{Log-Odds} of success, and DARS-HW optimizes the \textbf{Maximum-Likelihood} objective \(\log p\), where \(p=\text{pass@1}(q)\) denotes the prompt-level success rate. Table~\ref{tab:dars_objectives} summarizes the corresponding adaptive rollout budget, expected prompt-level gradient, and induced objective. The full derivation is deferred to Appendix~\ref{apd:implicit_obj}.

\begin{table}[t]
    \centering
    \small
    \renewcommand\arraystretch{1.6}
    \setlength{\tabcolsep}{3pt}
    \caption{Prompt-level objectives induced by different rollout allocation rules, where \(p=\text{pass@1}(q)\) and \(\mathbb{E}[\mathbf{g}_q]\) denotes the expected gradient contribution of prompt \(q\). \(C_{ET}=\frac{N^{pre}}{4}\) and \(C_{HW}=\frac{N^{pre}}{2}\) are constant values.}
    \begin{tabular}{l|c|c|c}
       \toprule
       Method & \(N(p)\) & \(\mathbb{E}[\mathbf{g}_q]\) & Training Objective \\
       \midrule
       \textbf{Dr. GRPO} & Fixed \(N\) & \(N \cdot \nabla_\theta p\) & \textbf{Pass@1} \\
       \textbf{DARS-ET} & \(\frac{N^{pre}}{4p(1-p)}\) & \(C_{ET}\cdot \nabla_\theta \log \frac{p}{1-p}\) & \textbf{Log-Odds}\\
       \textbf{DARS-HW} & \(\frac{N^{pre}}{2p}\) & \(C_{HW}\cdot \nabla_\theta \log p\) & \textbf{Max Likelihood}\\
       \bottomrule
    \end{tabular}
    \label{tab:dars_objectives}
\end{table}

\subsection{Depth Synergy with Breadth Scaling}
\label{sec:dars-b}
Our analysis in Section \ref{sec:analysis-breadth} empirically confirms the substantial \textit{Pass@1} improvements from large-breadth training. While DARS primarily optimizes training depth via multi-stage rollout rebalance, its dynamic batch-size adjustments preclude standard PPO-style mini-batch updates. To resolve this architectural constraint while leveraging breadth benefits, we replace PPO’s mini-batch updates with full-batch gradient descent across multiple PPO epochs, as illustrated in Figure \ref{fig:framework}. This modification ensures compatibility with DARS’s dynamic allocation while maximizing effective training breadth per optimization step. We term this integrated approach \textbf{DARS-Breadth}, unifying depth-adaptive sampling with breadth maximization.  

Full-batch training offers two key advantages: (1) elimination of mini-batch gradient noise, and (2) sustained token-level exploration, acting as implicit regularization against premature convergence. Empirically, DARS improves \textit{Pass@K} through depth optimization, while large-breadth training enhances \textit{Pass@1}, highlighting their synergistic roles in RLVR.

\begin{table*}[t]
    \vspace{3mm}
    \centering
    % \small
    % \scriptsize
    \renewcommand\arraystretch{1.3}
    \setlength{\tabcolsep}{4pt}
    \caption{Overall performance of \textit{Pass@1} (\textit{Avg@128}) and \textit{Pass@128} of Qwen2.5-Math series.}
    % \vspace{-2mm}
    \begin{tabular}{l|c|c|c|c|c|c|c}
       \toprule
       \textbf{Method}  & AIME24 & MATH500 & Olympiad & AMC & Minerva & \textbf{\textit{Avg@128}} & \textbf{\textit{Pass@128}}\\
       \midrule
       \rowcolor{gray!10} \multicolumn{8}{c}{\textit{Qwen2.5-Math-\highlight{1.5}B}} \\
        Base Modle & 4.0 & 35.1 & 16.2 & 20.8 & 9.5 & 21.1 & 77.9 \\
        RLVR baseline & 14.7 & 75.9 & 39.4 & 47.5 & 31.2 & 49.6 & 79.6 \\
        Depth-Naive & 16.5 & 76.2 & 39.9 & 47.9 & 30.9 & 50.1 & 79.9\\
        Breadth-Naive & 18.5 & 77.6 & 41.7 & 49.8 & \textbf{31.9} & 51.5 & 79.2 \\
        \rowcolor{darkgreen!8}\textbf{DARS-1.5B-ET} & 15.8 & 76.0 & 40.9 & 47.2 & 30.0 & 50.0 & 81.2\\
        \rowcolor{darkgreen!16}\textbf{DARS-1.5B-ET-Breadth} & \underline{18.6} & \textbf{79.4} & \textbf{42.9} & \underline{50.6} & \underline{31.7} & \textbf{52.5} & 80.8\\
        \rowcolor{table-blue!22}\textbf{DARS-1.5B-HW} & 17.7 & 76.4 & 40.0 & 48.4 & 30.8 & 50.0 & \underline{82.1}\\		
        \rowcolor{table-blue!66}\textbf{DARS-1.5B-HW-Breadth} & \textbf{19.3} & \underline{79.0} & \underline{42.7} & \textbf{51.9} & 31.6 & \underline{52.4} & \textbf{82.2} \\
        \midrule
        \rowcolor{gray!10} \multicolumn{8}{c}{\textit{Qwen2.5-Math-\highlight{7}B}} \\
        Base Model & 11.6 & 52.3 & 19.7 & 35.2 & 15.3 & 30.1 & 82.1 \\
        RLVR baseline & 26.8 & 82.2 & 44.3 & 57.2 & 35.7 & 55.3 & 81.4 \\
        Depth-Naive & 28.0 & 83.8 & 46.4 & 59.0 & 37.3 & 57.0 & 80.3 \\
        Breadth-Naive & 30.5 & 83.7 & 47.3 & \underline{61.4} & 37.7 & 57.7 & 79.2 \\
        \rowcolor{darkgreen!8}\textbf{DARS-7B-ET} & 26.9 & 83.2 & 46.6 & 57.3 & \textbf{38.5} & 57.0 & 81.7 \\
        \rowcolor{darkgreen!16}\textbf{DARS-7B-ET-Breadth} & \textbf{33.3} & \underline{83.8} & \underline{47.8} & 61.3 & \underline{38.4} & \underline{58.1} & 82.1\\
        \rowcolor{table-blue!22}\textbf{DARS-7B-HW} & 30.1 & 83.5 & 47.1 & 59.4 & 37.2 & 57.3 & \textbf{83.5} \\
        \rowcolor{table-blue!66}\textbf{DARS-7B-HW-Breadth}& \underline{33.0} & \textbf{84.5} & \textbf{48.4} & \textbf{63.0} & 36.9 & \textbf{58.4} & \underline{83.4} \\
        \midrule
        \rowcolor{gray!10} \multicolumn{8}{c}{\textit{Llama-3.1-\highlight{8B}}} \\
        Base Model & 0.23 & 6.13 & 1.54 & 2.76 & 2.72 & 3.25 & 52.7 \\
        GRPO baseline & 0.66 & 29.6 & 7.09 & 10.1 & 15.7 & 15.8 & 56.5 \\
        Depth-Naive & 0.43 & 33.6 & 9.40 & 12.3 & 19.7 & 18.9 & 58.6 \\
        Breadth-Naive & 0.79 & 34.4 & 9.34 & 12.2 & 19.0 & 19.0 & 61.1 \\
        \rowcolor{darkgreen!16}\textbf{DARS-Llama-ET-Breadth} & \textbf{1.46} & \textbf{39.4} & \textbf{12.0} & \underline{13.2} & \textbf{20.1} & \textbf{22.0} & 67.2 \\
        \rowcolor{table-blue!66}\textbf{DARS-Llama-HW-Breadth} & \underline{1.11} & \underline{39.0} & \textbf{12.0} & \textbf{13.3} & \underline{19.8} & \underline{21.8} & \textbf{68.7} \\
        \toprule
    \end{tabular}
    \label{tab:bench_scores}
    % \vspace{-6mm}
\end{table*}

\subsection{Training Target}
We adopt the clipped objective of GRPO without the KL penalty term. Following Dr.~GRPO, we likewise remove the response length handling from the GRPO target. Specifically, for a problem $q$ sampled in training data $\mathcal{D}$, the training target is formalized as:

\begin{footnotesize}
\begin{equation}
\begin{aligned}
&\mathcal{J}(\theta) = \mathbb{E}_{(q\sim \mathcal{D}, \{o_i\}_{i=1}^\mathcal{G}\sim \pi_{\theta_\text{old}}(q)}
\Bigg[ \frac{1}{G}\sum_{i=1}^{G}\sum_{t=1}^{|o_i|} \Bigg( 
\\&~~~~~\min \Big( r_{i,t}(\theta) \hat{A}_{i,t}, 
\ \text{clip} \Big( r_{i,t}(\theta), 1 - \varepsilon, 1 + \varepsilon \Big) \hat{A}_{i,t} \Big)
\Bigg) \Bigg],
\label{eq:grpoloss}
\end{aligned}
\end{equation}
\end{footnotesize}

where
\begin{equation}
    r_{i,t}(\theta)=\frac{\pi_{\theta}(o_{i,t} \mid q, o_{i,<t})}{\pi_{\theta_{\text{old}}}(o_{i,t} \mid q,o_{i,<t})}.
\end{equation}

The token advantage $\hat{A}_{i,t}$ is computed using Equation \ref{eq:grpo_adv}.

% \vspace{-3mm}
\begin{figure*}[htbp] % htbp    % width=0.99\textwidth
    \vspace{3mm}
    \centering
    \includegraphics[width=0.99\linewidth]{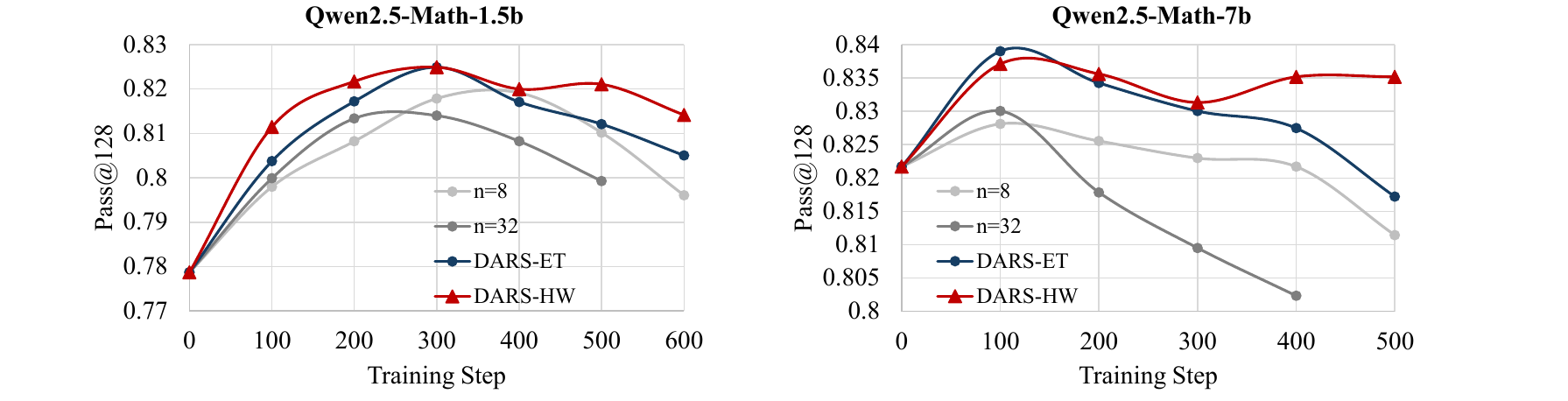}
    % \vspace{-8mm}
    % \vspace{-2mm}
    \caption{Training dynamics of \textit{Pass@128} performance with different training steps of Qwen2.5-Math-1.5B and Qwen2.5-Math-7B.}\label{fig:passkpeak}
    \vspace{3mm}
\end{figure*}

\begin{figure*}[t] % htbp    % width=0.99\textwidth
    \centering
    % \vspace{-4mm}
    \includegraphics[width=1.0\linewidth]{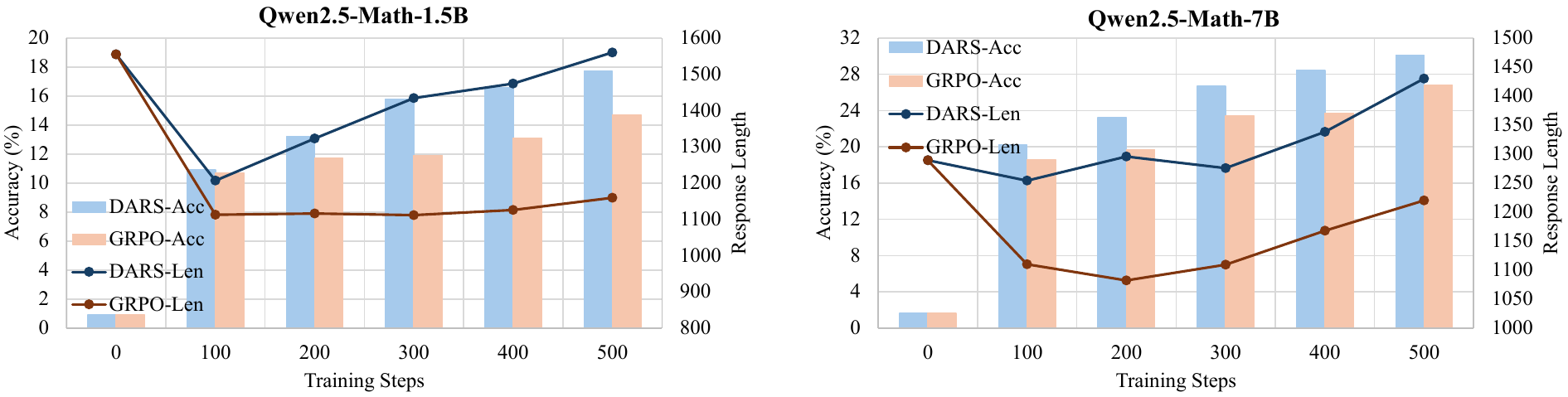}
    % \vspace{-6mm}
    \caption{Comparison of our DARS-HW and Dr. GRPO Baseline on AIME 2024. We show the average response length and accuracy for Qwen2.5-Math-1.5B and Qwen2.5-Math-7B. After training with DARS, the models can achieve higher accuracy by utilizing longer thinking process.}\label{fig:aime-length-acc}
    \vspace{1mm}
\end{figure*}

% \vspace{-5mm}
\section{Experiments}
\subsection{Setup}
\label{sec:setting}
\textbf{Evaluation and Training:} We evaluate the RLVR process using 5 widely used mathematical reasoning benchmarks: MATH-500~\citep{verify_stepbystep}, OlympiadBench~\citep{olympiadbench}, MinvervaMath~\citep{minerva}, AIME24, and AMC23. We combine all of the evaluation benchmarks to report \textit{Pass@1} (\textit{Avg@128}) and \textit{Pass@K} performance.
The training data used in this work is OpenR1-45K, which is a subset of OpenR1-Math-220k~\citep{openr1}. 
Moreover, we adopt the same unbiased, low-variance estimator for \textit{Pass@K} as used in prior works~\citep{yue2025does, chen2021codex}:
% \vspace{-1mm}
$$
\text{pass@}K = \mathbb{E}_{x_i \sim \mathcal{D}} \left[ 1 - \frac{\binom{n - c_i}{k}}{\binom{n}{k}} \right].
$$
% \vspace{-2mm}
Specifically, when $K=N$, the metric become: $\text{pass@}K = c_1 \lor c_2 \lor \cdots c_{128}$.

% \vspace{-1mm}
% \vspace{4mm}
\noindent
\textbf{Baseline and Methods:}
We compare with:
(1) \textit{RLVR-baseline}: Dr.~GRPO with rollout size 8 and batch size 128.  
(2) \textit{Depth-Naive}: Simply increasing the rollout size to 32.  
(3) \textit{Breadth-Naive}: Simply increasing the batch size to 3072.  
(4) \textit{DARS-ET/HW}: Our algorithm introduced in Section \ref{sec:dars} with Equal-Treat/Hardness-Weighted schedule, using batch size 128 and \( N^{\max}=32 \).  
(5) \textit{DARS-ET/HW-Breadth}: Our Depth-and-Breadth synergy algorithm introduced in Section \ref{sec:dars-b}, using batch size 3072 and \( N^{\max}=32 \).
For all methods, the number of PPO mini-steps is uniformly set to 2.

% \vspace{4mm}
\noindent
\textbf{Evaluation Protocol:}
For all baselines, we select the checkpoint with the best \textit{Pass@1} performance for reporting. For DARS, we selected the checkpoint that achieved the best\textit{ Pass@128} performance after surpassing the baseline \textit{Pass@1} performance.
Table \ref{tab:bench_scores} summarizes the \textit{Avg@128} performance on each benchmark, the overall \textit{Pass@1} across all test data, and the \textit{Pass@128} performance.

\subsection{Main Results}

Breadth scaling delivers a clear and consistent boost to Pass@1. Across every model scale and every benchmark, Breadth-Naive outperforms both the RLVR baseline and Depth-Naive, lifting average Pass@1 (Avg@128) by 1.9–3.7 points on AIME24, MATH500, and Olympiad tasks. This advantage is not merely additive: when depth is combined with breadth through DARS-Breadth, the margin widens further. DARS-Breadth reliably beats both Breadth-Naive and the original DARS variants, confirming our central hypothesis: depth and breadth are complementary, not competing resources. 

The practical impact is twofold. First, DARS-Breadth secures the highest Pass@1, the metric that matters most for single-shot deployment. Second, it matches the best Pass@128 scores, demonstrating that the breadth-depth collaboration does not sacrifice the upper-bound capability revealed by heavy sampling. Finally, the choice of schedule matters: the HW schedule consistently yields superior Pass@K curves for both breadth and non-breadth training, while maintaining Pass@1 parity with the ET schedule, making it the preferred option across the board.

% \vspace{-2mm}

\begin{table*}[t]
    \centering
    \footnotesize
    \renewcommand\arraystretch{1.2}
    \setlength{\tabcolsep}{4pt}
    \caption{Benchmark-level test-time scaling performance of the best checkpoints. \textit{maj@k} is averaged over five independent sampling rounds.}
    \resizebox{0.98\linewidth}{!}{
    \begin{tabular}{lccccccccc}
        \toprule
        \textbf{Method} & \multicolumn{3}{c}{\textbf{AIME24}} & \multicolumn{3}{c}{\textbf{AMC23}} & \multicolumn{3}{c}{\textbf{MinervaMath}} \\
        \cmidrule(lr){2-4} \cmidrule(lr){5-7} \cmidrule(lr){8-10}
        & \textit{maj@8} & \textit{maj@16} & \textit{pass@128} & \textit{maj@8} & \textit{maj@16} & \textit{pass@128} & \textit{maj@8} & \textit{maj@16} & \textit{pass@128} \\
        \midrule
        \rowcolor{gray!10} \multicolumn{10}{c}{\textit{Qwen2.5-Math-\highlight{1.5}B}} \\
        RLVR Baseline & 19.3 & 22.7 & 56.7 & 53.7 & 56.8 & 91.6 & 33.9 & 35.4 & 58.8 \\
        \textbf{DARS-HW} & 18.7 & 22.7 & 60.0 & 55.2 & 58.8 & \textbf{94.0} & \textbf{34.9} & \textbf{36.1} & 58.8 \\
        \textbf{DARS-HW-Breadth} & \textbf{22.7} ({\color{darkgreen}$\uparrow$3.4}) & \textbf{27.3} ({\color{darkgreen}$\uparrow$4.6}) & \textbf{66.7} ({\color{darkgreen}$\uparrow$10.0}) & \textbf{58.6} ({\color{darkgreen}$\uparrow$4.9}) & \textbf{59.5} ({\color{darkgreen}$\uparrow$2.7}) & 93.9 ({\color{darkgreen}$\uparrow$2.3}) & 34.6 ({\color{darkgreen}$\uparrow$0.7}) & 35.4 & \textbf{61.0} ({\color{darkgreen}$\uparrow$2.2}) \\
        \midrule
        \rowcolor{gray!10} \multicolumn{10}{c}{\textit{Qwen2.5-Math-\highlight{7}B}} \\
        RLVR Baseline & 27.3 & 33.3 & 73.3 & 63.1 & 64.3 & 91.6 & 38.5 & \textbf{39.6} & 58.5 \\
        \textbf{DARS-HW} & 37.3 & 43.3 & \textbf{76.7} & 68.0 & 69.6 & 95.2 & 38.6 & 39.3 & 62.1 \\
        \textbf{DARS-HW-Breadth} & \textbf{43.3} ({\color{darkgreen}$\uparrow$16.0}) & \textbf{44.7} ({\color{darkgreen}$\uparrow$11.4}) & \textbf{76.7} ({\color{darkgreen}$\uparrow$3.4}) & \textbf{68.4} ({\color{darkgreen}$\uparrow$5.3}) & \textbf{69.9} ({\color{darkgreen}$\uparrow$5.6}) & \textbf{97.6} ({\color{darkgreen}$\uparrow$6.0}) & \textbf{39.0} ({\color{darkgreen}$\uparrow$0.5}) & 39.4 ({\color{red!30}$\downarrow$0.2}) & \textbf{64.3} ({\color{darkgreen}$\uparrow$5.8}) \\
        \bottomrule
    \end{tabular}
    }
    \label{tab:test_time_scaling}
\end{table*}

%%%% 表格文字不并行

\subsection{Training Dynamics}
% \vspace{-1mm}
In this subsection, we further show more training dynamics to illustrate properties of existing RLVR methods and the superiority of our DARS and DARS-B.

\noindent
\textbf{\textit{Pass@128} performance surpasses the base model, peaks quickly, and then declines.} We conduct RLVR experiments with rollout size 8/32 to compare our DARS (with $N^{\mathrm{max}}=32$), the training dynamics of \textit{Pass@128} performance during training is shown in Figure \ref{fig:passkpeak}. Across all settings, \textit{Pass@128} surpasses the base model during training, but declines after peaking, indicating that over-training with RLVR harms \textit{Pass@128} performance. 
Notably, DARS (with $N^{\mathrm{max}}=32$) incurs substantially less inference cost than naively scaling the rollout size to n = 32. Despite this being an unfair comparison in terms of computational expenditure, our DARS not only attains the highest peak \textit{Pass@128} performance but also outperforms all other settings.

\noindent
\textbf{Depth and Breadth Are Complementary in RLVR.}
\label{sec:breadth-method}
We show that Depth and Breadth are two complementary dimensions in RLVR, with breadth primarily governing \textit{Pass@1} and depth primarily governing \textit{Pass@K}. 
As shown in Figure \ref{fig:D_B_128}, we present the \textit{Pass@1}–\textit{Pass@K} training-dynamics curves for the Breadth, Depth, and the two-dimensional synergy approach DARS-Breadth. The farther the \textit{Pass@1}–\textit{Pass@K} curve deviates outward, the more powerful the method. Our DARS-Breadth curves lie on the outermost envelope: it not only achieves the best \textit{Pass@1}, but also simultaneously lifts \textit{Pass@K}. This demonstrates the complementary roles of Depth and Breadth.

\noindent
\textbf{DARS Elicits Longer Thought.}
We tracked the response length dynamics during the training of DARS and Dr.GRPO. As shown in Figure \ref{fig:aime-length-acc}, on the challenging AIME 2024 benchmark, DARS consistently produces longer reasoning traces than the baseline and has higher accuracy. These results reveal that our DARS stimulates the model to perform deeper thinking to solve hard problems.

\subsection{Test-Time Scaling Performance}
We further evaluate whether DARS continues to benefit from additional test-time compute. Specifically, we compare the best checkpoints using majority voting (\textit{maj@8}, \textit{maj@16}) and \textit{pass@128} on AIME24, AMC23, and MinervaMath. The \textit{maj@k} results are averaged over five independent rounds of random sampling. We report this benchmark-level breakdown in Table \ref{tab:test_time_scaling}, since the overall \textit{pass@128} in Table \ref{tab:bench_scores} is aggregated across all evaluation instances and can partially mask gains on smaller but challenging benchmarks such as AIME24 and AMC23.

Table \ref{tab:test_time_scaling} shows that DARS consistently improves test-time scaling performance. For Qwen2.5-Math-1.5B, DARS-HW-Breadth improves the GRPO baseline by 4.6 points on AIME24 \textit{maj@16}, 10.0 points on AIME24 \textit{pass@128}, 2.7 points on AMC23 \textit{maj@16}, and 2.2 points on MinervaMath \textit{pass@128}. The gains are even larger on Qwen2.5-Math-7B, where DARS-HW-Breadth improves AIME24 \textit{maj@8}/\textit{maj@16} by 16.0/11.4 points, while also raising AMC23 \textit{pass@128} and MinervaMath \textit{pass@128} by 6.0 and 5.8 points, respectively. These results show that DARS not only improves single-sample behavior but also yields a stronger solution space under test-time search.

\begin{table}[t]
    \renewcommand\arraystretch{1.3}
  \centering
  \footnotesize
  % \vspace{-6mm}
  \caption{Average rollout numbers per prompt and wall-clock time per training step (seconds).}
  % \vspace{-2mm}
  \resizebox{0.98\linewidth}{!}{
  \begin{tabular}{lcc}
    \toprule
    Method & Avg. Rollouts / Prompt & Time / Step (s) \\ 
    \midrule
    \rowcolor{gray!10} \multicolumn{3}{c}{\textit{Qwen2.5-Math-\highlight{1.5}B}} \\
    Depth-Naive & 32 & 179 \\
    DARS-ET & 15.2 ({\color{darkgreen!100}$\downarrow$\textbf{52.5}\%}) & 115 ({\color{darkgreen!100}$\downarrow$\textbf{35.8}\%}) \\
    DARS-HW & 23.9 ({\color{darkgreen!100}$\downarrow$\textbf{25.3}\%}) & 160 ({\color{darkgreen!100}$\downarrow$\textbf{10.6}\%}) \\
    \midrule
    \rowcolor{gray!10} \multicolumn{3}{c}{\textit{Qwen2.5-Math-\highlight{7}B}} \\
    Depth-Naive & 32 & 263 \\
    DARS-ET & 12.8 ({\color{darkgreen!100}$\downarrow$\textbf{60.0}\%}) & 174 ({\color{darkgreen!100}$\downarrow$\textbf{33.8}\%}) \\
    DARS-HW & 20.1 ({\color{darkgreen!100}$\downarrow$\textbf{37.2}\%}) & 226 ({\color{darkgreen!100}$\downarrow$\textbf{14.1}\%}) \\
    \bottomrule
  \end{tabular}
  }
  \label{tab:efficiency}
  \vspace{-2mm}
\end{table}

\subsection{Ablation Studies}
\noindent
\textbf{Ablation Study on Base Model.}
We illustrate the effectiveness of DARS on different models. The results are shown in Table \ref{tab:bench_scores}. Our DARS-ET-Breadth achieves both the highest \textit{Pass@1} and \textit{Pass@128} performance, which further illustrates the effectiveness of our method. DARS also consistently improves the performance of openPangu \cite{chen2025pangu}, as shown in Appendix \ref{apd:more_exp}. We further show the comparison results on static DARS in Appendix \ref{apd:static_dars}.

\subsection{Impact of Temperature}

\begin{figure*}[t] % htbp    % width=0.99\textwidth
    \centering
    \includegraphics[width=0.95\linewidth]{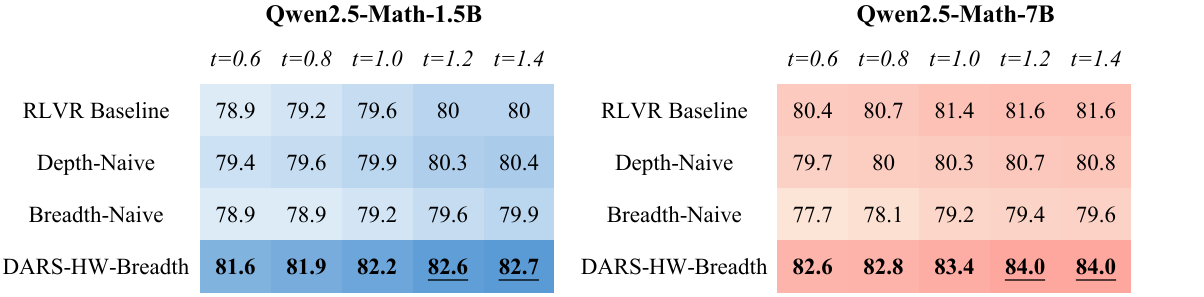}
    % \vspace{-1mm}
    \caption{Heat map of \textit{Pass@128} of Qwen2.5-Math series in different temperatures.}
    \label{fig:heat_map}
    % \vspace{1mm}
\end{figure*}

Some researches \citep{karan2025reasoningsamplingbasemodel,qin2025decomposingelementsproblemsolving,ni2025grpohelpllmstranscend} indicates that temperature matters in LLM reasoning. To further illustrate the performance improvement under different temperatures, we additionally add the above experiments. The results show that the improvement of our method is consistent over different temperatures. The results are shown in Figure \ref{fig:heat_map}.

\begin{figure*}[t] % htbp    % width=0.99\textwidth
    \centering
    \includegraphics[width=1.0\linewidth]{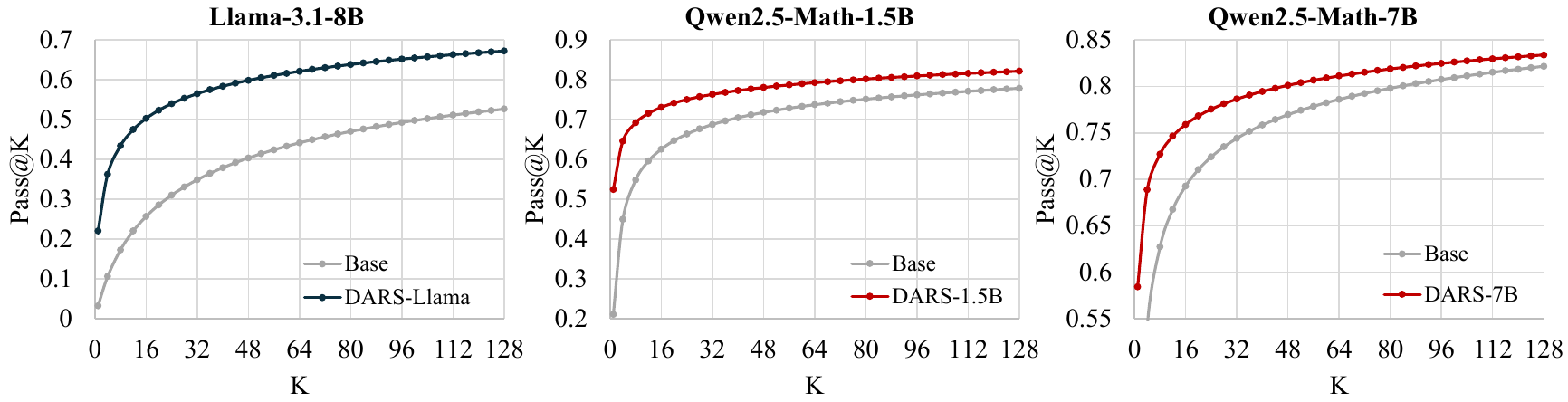}
    \vspace{-4mm}
    \caption{Complete \textit{Pass@K} accuracy curve of base models and our DARS models.}
    \label{fig:complete_passk}
    % \vspace{-1mm}
\end{figure*}

% \noindent
\subsection{DARS Improve \textit{Pass@K} and Training Efficiency}
Because the \textit{Pass@K} (K=32/128) metric is hard to improve monotonically: it even starts to drop after prolonged training, while \textit{Pass@1} remains comparatively stable and rarely collapses. We seek to boost \textit{Pass@K} without degrading \textit{Pass@1}. Figure \ref{fig:depth-capacity} plots \textit{Pass@128} against \textit{Pass@1} under a variety of experimental settings. It shows that, at any fixed \textit{Pass@1} level, our DARS method delivers a consistently higher \textit{Pass@128} than the other settings. 

It is worth noting that, unlike the naive approach of simply increasing the rollout size to 32, our DARS achieves significantly higher training efficiency by allocating more rollouts to the hard problems. As shown in Table \ref{tab:efficiency}, our DARS methods need far fewer rollouts than the Depth-Naive method while achieving better performance.

\subsection{Complete \textit{Pass@K} Accuracy Curve.}
We show the complete \textit{Pass@K} curve for Llama-3.1-8B, Qwen2.5-Math-1.5B, and Qwen2.5-Math-7B in Figure \ref{fig:complete_passk}. 
The 3 chosen models of DARS are: DARS-Llama-ET-Breadth, DARS-1.5B-HW-Breadth, and DARS-7B-HW-Breadth.
DARS models demonstrate a breakthrough in the reasoning boundaries of the base model, especially on the LLama-3.1-8B model, where the improvement in Pass@k is particularly significant.

\section{Related Works}
Reinforcement Learning (RL) is now standard in post-training LLMs. After early reward-model pipelines \citep{ouyang2022training}, Direct Preference Optimization \citep{rafailov2023dpo} streamlined training by exploiting pairwise preferences. RL with verifiable rewards (RLVR) has since pushed reasoning benchmarks in math and code, culminating in OpenAI’s o1 \citep{jaech2024openaio1} and the zero-RL breakthrough of DeepSeek-R1 \citep{guo2025deepseek-r1}. Follow-up Large Reasoning Models: Kimi 1.5 \citep{team2025kimi}, Gemini-Think \citep{gemini-thinking}, QwQ \citep{qwq}, and studies like \cite{zeng2025simplerl,deepscaler2025} further validate RLVR. The leading algorithm, GRPO \citep{deepseekmathgrpo}, extends PPO \citep{schulman2017proximal} with group-relative advantages, inspiring DAPO \citep{dapo}, VAPO \citep{vapo}, and Dr. GRPO \citep{liu2025understanding}. Yet GRPO and its variants systematically undervalue hard problems, hurting Pass@K. Inspired by Focal Loss \citep{lin2017focal}, this paper proposes DARS to explicitly up-weight the gradient contribution of challenging instances, ensuring the model explores sufficiently deep reasoning paths to unlock its potential.
More related works are shown in Appendix \ref{apd:related_works}.

% \vspace{-2mm}

% \subsection{Analysis of RLVR}
% With the rapid advancement of RLVR and the proliferation of open-source LRMs, many studies have begun to analyze the RLVR pipeline and these open LRMs. Several studies~\citep{liu2025there,zhao2025echo,shah2025rethinking} indicates that the self-reflect and self-critique behaviors observed after RLVR originates from the base model rather than the RL process. \cite{dang2025assessing} find that although the RLVR process benefits \textit{Pass@1}, \textit{Pass@K} may decline as training progresses.
% Subsequently, \cite{yue2025does} through extensive experimental analysis, discovered that RLVR’s performance is significantly constrained by the base model; once training converges, it struggles to surpass the capability boundary of the base model. These studies have sparked widespread concern about the capability ceiling of RLVR, and consequently, the \textit{Pass@K} metric has become a focal point for diagnosing and potentially transcending the intrinsic limits imposed by the base model~\citep{liang2025beyond}. This paper analyzes and refines the RLVR pipeline from the dual perspectives of \textit{Pass@1} and \textit{Pass@K}. 

\vspace{-2mm}
\section{Conclusion}
% \vspace{-1mm}
% In this work, we identify and systematically analyze two critical yet under-explored dimensions in Reinforcement Learning with Verifiable Reward (RLVR): \textbf{Depth} (the hardest problems a model can learn) and \textbf{Breadth} (the number of instances processed per iteration). We reveal that existing GRPO-based methods suffer from a fundamental limitation—their cumulative advantage calculation disproportionately weights medium-difficulty problems while neglecting hard ones, creating a bottleneck for \textit{Pass@K} performance. To address this issue, we propose Difficulty-Adaptive Rollout Sampling (\textbf{DARS}) to reallocate compute to the hard problems, boosting \textit{Pass@K} with low extra inference cost. Furthermore, we demonstrate that large-breadth training acts as an orthogonal dimension for RLVR optimization. Scaling batch size and employing full-batch updates significantly boost \textit{Pass@1} performance while sustaining high token-level entropy.  By unifying these dimensions in our DARS-Breadth framework, we achieve simultaneous gains in both \textit{Pass@1} and \textit{Pass@K}, confirming the synergistic relationship between depth and breadth in RLVR.

In this work, we reveal that GRPO-based RLVR methods under-weight hard problems due to cumulative-advantage bias, capping \textit{Pass@K}. Our DARS sampler efficiently re-allocates rollouts to these hard instances, while large-breadth training with full-batch updates raises \textit{Pass@1}. The unified DARS-Breadth jointly lifts \textit{Pass@1} and \textit{Pass@K}, proving depth and breadth are synergistic levers in RLVR.

\newpage

\section*{Impact Statement}

This paper presents work whose goal is to advance the field of Machine
Learning. There are many potential societal consequences of our work, none
which we feel must be specifically highlighted here.


% \bibliography{example_paper}
\bibliographystyle{icml2026}

%%%%%%%%%%%%%%%%%%%%%%%%%%%%%%%%%%%%%%%%%%%%%%%%%%%%%%%%%%%%%%%%%%%%%%%%%%%%%%%
%%%%%%%%%%%%%%%%%%%%%%%%%%%%%%%%%%%%%%%%%%%%%%%%%%%%%%%%%%%%%%%%%%%%%%%%%%%%%%%
% APPENDIX
%%%%%%%%%%%%%%%%%%%%%%%%%%%%%%%%%%%%%%%%%%%%%%%%%%%%%%%%%%%%%%%%%%%%%%%%%%%%%%%
%%%%%%%%%%%%%%%%%%%%%%%%%%%%%%%%%%%%%%%%%%%%%%%%%%%%%%%%%%%%%%%%%%%%%%%%%%%%%%%
\newpage
\appendix
\onecolumn
\section*{Appendix}

\section*{Contents of Appendix}
\vspace{-0.4em}
% \noindent{\color{black!35}\rule{\linewidth}{0.5pt}}
\vspace{0.25em}
\begingroup
% \small
\setlength{\parindent}{0pt}
\setlength{\parskip}{0pt}
\appcontentsline{apd:related_works}{More Related Works}
\appcontentsline{app:math_deriv}{Mathematical Derivations for DARS and GRPO Training Dynamics}
\appsubcontentsline{app:delta_rollouts}{Derivation of Additional Rollouts $\Delta n_j$}
\appsubcontentsline{apd:implicit_obj}{Implicit Optimization Objectives of the ET and HW Reallocation Rules}
\appcontentsline{apd:compare_dars_maxrl}{Theoretical Connection Between DARS and Maximum Likelihood Reinforcement Learning}
\appsubcontentsline{app:objective_equiv}{Objective Equivalence via Cumulative Advantage}
\appsubcontentsline{app:expected_grad_equiv}{Equivalence of Expected Gradients}
\appsubcontentsline{apd:variance}{Variance Analysis: DARS-HW versus MaxRL}
\appsubcontentsline{app:theory_conclusion}{Theoretical Analysis Conclusions}
\appcontentsline{apd:detail}{Training and Evaluation Details}
\appcontentsline{apd:more_exp}{More Experimental Results}
\appsubcontentsline{apd:ood_domain}{Generalization to Other Task Domains}
\appsubcontentsline{app:std_ablation}{Ablation Study on std-based Advantage Computation}
\appsubcontentsline{apd:static_dars}{Comparison with Static Difficulty Allocation}
\appsubcontentsline{app:training_dynamics_compare}{Training Dynamics Comparison of DARS-ET/HW and Depth-Naive}
\appsubcontentsline{apd:pass32-1}{Depth and Breadth Synergy for Pass@1 and Pass@32}
\appsubcontentsline{app:et_hw_breadth}{Comparison of ET/HW Schedule in Breadth Scaling}
\appsubcontentsline{app:consistent_improve_rl}{Consistent Improvement During RL Process}
\appsubcontentsline{app:other_model}{Performance of Other Model}
\appsubcontentsline{app:llama3_passk}{Explanation of Significant Pass@k Gain for Llama-3}
\appsubcontentsline{app:thinking_length}{Thinking Length Dynamics}
\appcontentsline{app:case_study}{Case Study}
\appcontentsline{app:discussion_future}{Discussion and Future Work}
\appsubcontentsline{apd:hyperparam}{Hyperparameter Control of Cumulative Advantage Shape}
\appsubcontentsline{app:passk_to_pass1_transition}{Potential Pass@K to Pass@1 Training Transition}
\endgroup
\vspace{0.2em}
\noindent{\color{black!35}\rule{\linewidth}{0.5pt}}
\vspace{0.6em}

\section{More Related Works}
\label{apd:related_works}
With the rapid advancement of RLVR and the proliferation of open-source LRMs, many studies have begun to analyze the RLVR pipeline and these open LRMs. Several studies~\citep{liu2025there,zhao2025echo,shah2025rethinking} indicates that the self-reflect and self-critique behaviors observed after RLVR originates from the base model rather than the RL process. \cite{dang2025assessing} find that although the RLVR process benefits \textit{Pass@1}, \textit{Pass@K} may decline as training progresses.
Subsequently, \cite{yue2025does} through extensive experimental analysis, discovered that RLVR’s performance is significantly constrained by the base model; once training converges, it struggles to surpass the capability boundary of the base model. These studies have sparked widespread concern about the capability ceiling of RLVR, and consequently, the \textit{Pass@K} metric has become a focal point for diagnosing and potentially transcending the intrinsic limits imposed by the base model~\citep{liang2025beyond}. 
This paper analyzes and refines the RLVR pipeline from the dual perspectives of \textit{Pass@1} and \textit{Pass@K}.

\section{Mathematical Derivations for DARS and GRPO Training Dynamics}
\label{app:math_deriv}
\subsection{Derivation of Additional Rollouts $\Delta n_j$}
\label{app:delta_rollouts}
The cumulative advantage for a group with accuracy $\hat{a}_j$ and total rollout size $N_j = {N^{pre}} + \Delta n_j$ is given by:
\begin{equation*}
\mathcal{A}_{\text{group}}(\hat{a}_j, N_j) = N_j \cdot \mathcal{S}(\hat{a}_j),
\end{equation*}
where $\mathcal{S}(\hat{a}_j) = 2\hat{a}_j(1-\hat{a}_j)$.

After the first-stage rollout of size ${N^{pre}}$, the initial cumulative advantage is:
\begin{equation*}
\mathcal{A}_{\text{group}}^{N^{pre}}(\hat{a}_j) = {N^{pre}} \cdot \mathcal{S}(\hat{a}_j).
\end{equation*}

Our goal is to determine the number of additional trajectories $\Delta n_j$ needed so that the final cumulative advantage $\mathcal{A}_{\text{group}}(\hat{a}_j, N_j)$ meets a target value $\mathcal{A}_{\text{group}}^{\text{target}}(\hat{a}_j)$.

\noindent
\textbf{Equal-Treatment (ET) Schedule:}

The target cumulative advantage is set to be constant for all questions with $\hat{a}_j < 0.5$:
\begin{equation*}
\mathcal{A}_{\text{group}}^{\text{ET}}(\hat{a}_j) = \mathcal{A}_{\text{group}}^{N^{pre}}(0.5) = {N^{pre}} \cdot \mathcal{S}(0.5).
\end{equation*}

We solve for $\Delta n_j^{\text{ET}}$:
\begin{align*}
\mathcal{A}_{\text{group}}(\hat{a}_j, N_j) &= \mathcal{A}_{\text{group}}^{\text{ET}}(\hat{a}_j) \\
({N^{pre}} + \Delta n_j^{\text{ET}}) \cdot \mathcal{S}(\hat{a}_j) &= {N^{pre}} \cdot \mathcal{S}(0.5) \\
\Delta n_j^{\text{ET}} \cdot \mathcal{S}(\hat{a}_j) &= {N^{pre}} \cdot \mathcal{S}(0.5) - {N^{pre}} \cdot \mathcal{S}(\hat{a}_j) \\
\Delta n_j^{\text{ET}} &= \frac{{N^{pre}} \cdot \mathcal{S}(0.5) - {N^{pre}} \cdot \mathcal{S}(\hat{a}_j)}{\mathcal{S}(\hat{a}_j)}. \\
\Delta n_j^{\text{ET}} &=  \frac{\mathcal{A}_{\text{group}}^{N^{pre}}(0.5) - \mathcal{A}_{\text{group}}^{N^{pre}}(\hat{a}_j)}{\mathcal{S}(\hat{a}_j)}.
\end{align*}

The rollout size must be an integer, and we cap the total rollout sampling upper limit at $N^{max}$, so
\begin{equation*}
\Delta n_j^{\text{ET}}=\mathrm{min}(
\left\lceil
\frac{\mathcal{A}_{\text{group}}^{N^{pre}}(0.5)-\mathcal{A}_{\text{group}}^{N^{pre}}(\hat{a}_j)}
{\mathcal{S}(\hat{a}_j)}
\right\rceil, N^{\mathrm{max}} - N^{pre}).
\end{equation*}

\noindent
\textbf{Hardness-Weighted (HW) Schedule:}

The target cumulative advantage increases with difficulty:
\begin{equation*}
\mathcal{A}_{\text{group}}^{\text{HW}}(\hat{a}_j) = 2(1 - \hat{a}_j) \cdot \mathcal{A}_{\text{group}}^N(0.5) = 2x_j \cdot N^{pre} \cdot \mathcal{S}(0.5).
\end{equation*}

We solve for $\Delta n_j^{\text{HW}}$:
\begin{align*}
\mathcal{A}_{\text{group}}(\hat{a}_j, N_j) &= \mathcal{A}_{\text{group}}^{\text{HW}}(\hat{a}_j) \\
(N^{pre} + \Delta n_j^{\text{HW}}) \cdot \mathcal{S}(\hat{a}_j) &= 2x_j \cdot N^{pre} \cdot \mathcal{S}(0.5) \\
\Delta n_j^{\text{HW}} \cdot \mathcal{S}(\hat{a}_j) &= 2x_j \cdot N^{pre} \cdot \mathcal{S}(0.5) - N^{pre} \cdot \mathcal{S}(\hat{a}_j) \\
\Delta n_j^{\text{HW}} &= \frac{2x_j \cdot N^{pre} \cdot \mathcal{S}(0.5) - N^{pre} \cdot \mathcal{S}(\hat{a}_j)}{\mathcal{S}(\hat{a}_j)}.
\end{align*}

Again, using the baseline advantage notation $\mathcal{A}_{\text{group}}^{N^{pre}}(\hat{a}_j) = N^{pre} \cdot \mathcal{S}(\hat{a}_j)$, we obtain:
\begin{equation*}
\Delta n_j^{\text{HW}}=\mathrm{min}(
\left\lceil
\frac{2x_j \cdot \mathcal{A}_{\text{group}}^{N^{pre}}(0.5)- \mathcal{A}_{\text{group}}^{N^{pre}}(\hat{a}_j)}
{\mathcal{S}(\hat{a}_j)}
\right\rceil, N^{\mathrm{max}} - N^{pre}).
\end{equation*}

\subsection{Implicit Optimization Objectives of the ET and HW Reallocation Rules}
\label{apd:implicit_obj}
We next analyze the prompt-level population objectives induced by the reallocation schedules under the Dr.~GRPO-style unnormalized group-relative advantage used in our main experiments. Let \(p \in (0,1)\) denote the true pass rate of a prompt \(q\), and let the difficulty score satisfy \(x=1-p\). In this subsection we use the idealized population approximation \(u \mapsto p\), replace the empirical estimate \(\hat{a}_j\) with its population value \(p\), and ignore the ceiling operator and the cap \(N^{\max}\). Under this approximation, define the surrogate prompt-level gradient
\begin{equation}
\mathbf{g}_q(N) \triangleq \sum_{i=1}^{N} (r_i-p)\nabla_\theta \log \pi_\theta(o_i|q),
\end{equation}
where \(r_i\in\{0,1\}\) is the rollout reward. We condition on a fixed prompt \(q\), assume the rollouts \(o_1,\dots,o_N\) are sampled i.i.d. from the current policy \(\pi_\theta(\cdot|q)\), and write expectations with respect to this rollout randomness. This population surrogate keeps the derivation concise; using the exact sample-mean baseline \(\bar r_N\) only changes constant factors by a finite-sample correction and does not alter the induced objective family.

\begin{proposition-block}[title=Proposition B.2 (Idealized Population Objectives of ET and HW under Dr.~GRPO)]
\vspace{-1.5mm}
In the idealized continuous analysis above, DARS-HW induces
\(
\mathbb{E}[\mathbf{g}^{\mathrm{HW}}_q]=\frac{N^{pre}}{2}\nabla_\theta \log p
\),
which is proportional to the \textbf{Maximum-Likelihood} gradient.
Likewise, DARS-ET induces
\(
\mathbb{E}[\mathbf{g}^{\mathrm{ET}}_q]=\frac{N^{pre}}{4}\nabla_\theta \log \frac{p}{1-p}
\),
which is proportional to the \textbf{Log-Odds} gradient.
\vspace{-1.5mm}
\end{proposition-block}

\vspace{-1mm}
\begin{proof}
For a single rollout, the score-function identity gives
\vspace{-1mm}
\begin{equation}
\begin{aligned}
\mathbb{E}[\mathbf{g}_q(N)]
&= \sum_{i=1}^{N}\mathbb{E}\left[(r_i-p)\nabla_\theta \log \pi_\theta(o_i|q)\right] \\
&= \sum_{i=1}^{N}\left(
\mathbb{E}\left[r_i\nabla_\theta \log \pi_\theta(o_i|q)\right]
- p\,\mathbb{E}\left[\nabla_\theta \log \pi_\theta(o_i|q)\right]
\right) \\
&= N \cdot \nabla_\theta p.
\end{aligned}
\label{eq:drgrpo_prompt_grad_surrogate}
\end{equation}\vspace{-1mm}
Under the same population approximation, the expected cumulative advantage is
\begin{equation}
\mathbb{E}[\mathcal{A}_{\text{group}}(N,p)]
 = 2Np(1-p).
\label{eq:drgrpo_group_adv_surrogate}
\end{equation}
The medium-difficulty reference generated by the first-stage pre-rollout is therefore
\begin{equation}
\mathcal{A}_{\mathrm{ref}}
= 2N^{pre}\cdot (0.5)(1-0.5)
= \frac{N^{pre}}{2}.
\end{equation}
\textbf{For the HW schedule}, Equation~\ref{eq:A_HW} sets the target cumulative advantage to
\begin{equation}
\mathcal{A}_{\text{target}}^{\mathrm{HW}}(p)
= 2(1-p)\mathcal{A}_{\mathrm{ref}}
= (1-p)N^{pre}.
\end{equation}
Equating this target with the expected group cumulative advantage yields
\begin{equation}
2N_{\mathrm{HW}}(p)p(1-p) = (1-p)N^{pre}
\quad \Longrightarrow \quad
N_{\mathrm{HW}}(p)=\frac{N^{pre}}{2p}.
\end{equation}\vspace{-1mm}
Substituting this rollout allocation into the expected gradient gives
\begin{equation}
\mathbb{E}[\mathbf{g}^{\mathrm{HW}}_q]
= N_{\mathrm{HW}}(p)\nabla_\theta p
= \frac{N^{pre}}{2p}\nabla_\theta p
= \frac{N^{pre}}{2} \nabla_\theta \log p.
\end{equation}
So, 
\begin{equation}
\mathbb{E}[\mathbf{g}^{\mathrm{HW}}_q]
= C_{HW}~~~\cdot\underbrace{\nabla_\theta \log p}_{\text{Maximum Likelihood}},
\end{equation}
where \(C_{HW}=\frac{N^{pre}}{2}\) is a constant.

\textbf{For the ET schedule}, Equation~\ref{eq:A_ET} fixes the target cumulative advantage to the same medium-difficulty anchor:
\begin{equation}
\mathcal{A}_{\text{target}}^{\mathrm{ET}}(p)
= \mathcal{A}_{\mathrm{ref}}
= \frac{N^{pre}}{2}.
\end{equation}
Equating this target with the expected group cumulative advantage yields
\begin{equation}
2N_{\mathrm{ET}}(p)p(1-p) = \frac{N^{pre}}{2}
\quad \Longrightarrow \quad
N_{\mathrm{ET}}(p)=\frac{N^{pre}}{4p(1-p)}.
\end{equation}
Therefore,
\begin{equation}
\mathbb{E}[\mathbf{g}^{\mathrm{ET}}_q]
= N_{\mathrm{ET}}(p)\nabla_\theta p
= \frac{N^{pre}}{4p(1-p)}\nabla_\theta p
= \frac{N^{pre}}{4} \nabla_\theta \log \frac{p}{1-p}.
\end{equation}

So, 
\begin{equation}
\mathbb{E}[\mathbf{g}^{\mathrm{ET}}_q]
= C_{ET}~~~\cdot\underbrace{\nabla_\theta \log \frac{p}{1-p}}_{\text{Log-Odds}},
\end{equation}
where \(C_{ET}=\frac{N^{pre}}{4}\) is a constant.

This proves the proposition.
\end{proof}

Proposition~B.2 reveals that the two schedules induce different population-level weighting functions under Dr.~GRPO: ET equalizes cumulative advantage and yields a \textbf{Log-Odds} weighting, whereas HW scales compute linearly with hardness and yields a \textbf{Maximum-Likelihood} weighting. The implemented algorithm further enforces integer rollout counts, a rollout cap \(N^{\max}\), and a two-stage pre-rollout floor. These implementation details discretize or truncate the idealized continuous objectives derived above, but do not change their functional form in this population analysis.

% \newpage
\newpage
\section{Theoretical Connection Between DARS and Maximum Likelihood Reinforcement Learning}
\label{apd:compare_dars_maxrl}
Recently, Maximum Likelihood Reinforcement Learning (MaxRL) \citep{maxrl} introduces a modified advantage function to approximate the Maximum Likelihood (ML) objective. 
Appendix \ref{apd:implicit_obj} already shows that the continuous uncapped HW schedule induces a gradient exactly proportional to \(\nabla_\theta \log p\). In this section, we further provide a formal comparison to MaxRL and demonstrate that \textbf{our Difficulty-Adaptive Rollout Sampling (DARS) with the Hardness-Weighted (HW) schedule optimizes an expected objective identical to MaxRL}.

\subsection{Objective Equivalence via Cumulative Advantage}
\label{app:objective_equiv}
We first analyze the weighting behavior of both algorithms utilizing our proposed Group Cumulative Advantage ($\mathcal{A}_{group}$) framework. Let $N$ denote the number of rollouts for a given prompt $q$, and let $u \in (0, 1)$ denote the mean reward (i.e., empirical accuracy or pass rate), where the binary reward is $r_i \in \{0, 1\}$.

\begin{proposition-block}[title=Proposition C.1 (Equivalence of Difficulty Weighting Profile)]
Both MaxRL and DARS-HW impose a problem-level optimization weight that is strictly proportional to $(1 - u)$. Consequently, both algorithms share an identical inductive bias: linearly up-weighting the optimization focus on problems as their difficulty increases ($u \to 0$).
\end{proposition-block}
\vspace{-2mm}

% \noindent\textbf{Proof of Proposition C.1.}
\begin{proof}
For MaxRL, the advantage is scaled as $\hat{A}^{MaxRL}_i = \frac{r_i - u}{u}$. For a group of $N$ rollouts, the expected number of successful trajectories ($r_i = 1$) is $Nu$, and the expected number of failed trajectories ($r_i = 0$) is $N(1-u)$. The cumulative advantage is thus computed as:
$$
\begin{aligned}
\mathcal{A}_{group}^{MaxRL}(u) &= \mathbb{E} \left[ \sum_{i=1}^N \left| \hat{A}^{MaxRL}_i \right| \right] \\
&= Nu \cdot \left| \frac{1 - u}{u} \right| + N(1 - u) \cdot \left| \frac{0 - u}{u} \right| \\
&= N(1-u) + N(1-u) = 2N(1 - u).
\end{aligned}
$$
For DARS-HW, we employ the unscaled advantage $\hat{A}_i = r_i - u$. Based on our proposed schedule (Equation 7), the target cumulative advantage is explicitly constrained to:
$$
\mathcal{A}_{group}^{HW}(u) = 2(1 - u) \mathcal{A}_{group}^{N_{pre}}(0.5) = C \cdot (1 - u).
$$
where $C = 2 \mathcal{A}_{group}^{N_{pre}}(0.5)$ is a constant independent of the prompt difficulty $u$. Since $\mathcal{A}_{group}^{MaxRL}(u) \propto (1-u)$ and $\mathcal{A}_{group}^{HW}(u) \propto (1-u)$, the weighting profiles are equivalent. 
% $\blacksquare$
\end{proof}

\subsection{Equivalence of Expected Gradients}
\label{app:expected_grad_equiv}
To formalize the alignment in optimization trajectories, we compare the total expected gradient contribution of a single prompt $q$ under both methods. Let $\nabla J_i(u) = (r_i - u) \nabla_\theta \log \pi_\theta(o_i | q)$ denote the standard baseline-subtracted policy gradient for a single rollout.

\begin{proposition-block}[title=Proposition C.2 (Equivalence of Expected Gradient Updates)]
Let $K$ be a positive scalar. The expected gradient updates of MaxRL and DARS-HW are proportional: $\mathbb{E}[\nabla J^{MaxRL}_q] \propto \mathbb{E}[\nabla J^{DARS-HW}_q] \propto \frac{1}{u} \mathbb{E} [\nabla J_i(u)]$. Both methods approximate the Maximum Likelihood gradient by universally scaling the standard policy gradient by a factor of $1/u$.
\end{proposition-block}

\begin{proof}
For MaxRL, using a fixed rollout size $N$, the total expected gradient is scaled algebraically:
\begin{equation*}
\mathbb{E}[\nabla J^{MaxRL}_q] = \mathbb{E} \left[ \sum_{i=1}^N \frac{r_i - u}{u} \nabla_\theta \log \pi_\theta(o_i | q) \right] = \frac{N}{u} \cdot \mathbb{E} \left[ \nabla J_i(u) \right].
\end{equation*}
For DARS-HW, we dynamically adjust the rollout size $N_{DARS}(u)$ to match the target $\mathcal{A}_{group}^{HW}(u) = C(1-u)$. Recall the standard group cumulative advantage property $\mathcal{A}_{group} = 2 N_{DARS} u (1-u)$. Equating the two yields $2 N_{DARS}(u) u (1-u) = C(1-u)$, which simplifies to $N_{DARS}(u) = \frac{C}{2u}$. Thus, the expected gradient contribution is:
\begin{equation*}
\mathbb{E}[\nabla J^{DARS-HW}_q] = \mathbb{E} \left[ \sum_{i=1}^{N_{DARS}(u)} (r_i - u) \nabla_\theta \log \pi_\theta(o_i | q) \right] = \frac{C}{2u} \cdot \mathbb{E} \left[ \nabla J_i(u) \right].
\end{equation*}
Up to the constants $N$ and $C/2$, both expected gradients scale strictly by $1/u$.
% $\blacksquare$
\end{proof}

\subsection{Variance Analysis: DARS-HW versus MaxRL}
\label{apd:variance}

While Proposition C.2 establishes that DARS-HW and MaxRL optimize the same expected objective, they diverge significantly in their construction of the Monte Carlo gradient estimator. Let $\Sigma_g = \text{Var}(\nabla J_i(u))$ denote the trace of the covariance matrix of the unscaled token-level gradient for a single rollout. We make the standard assumption that conditioned on the prompt $q$ and current policy $\pi_\theta$, individual rollouts are independent and identically distributed (i.i.d.).

Let $u \to 0$ denote the regime of increasingly difficult reasoning tasks. For the MaxRL estimator, the variance relies on algebraic scalar multiplication over a fixed $N$ samples. By the properties of variance for i.i.d. variables ($\text{Var}(aX) = a^2\text{Var}(X)$):
\begin{equation*}
\text{Var}(\hat{\nabla} J^{MaxRL}_q) = \text{Var} \left( \sum_{i=1}^N \frac{1}{u} \nabla J_i(u) \right) = \sum_{i=1}^N \frac{1}{u^2} \text{Var}(\nabla J_i(u)) = \frac{N}{u^2} \Sigma_g.
\end{equation*}
Conversely, DARS-HW physically expands the sampling space, drawing $N_{DARS}(u) = \frac{C}{2u}$ independent trajectories while maintaining the bounded unscaled advantage $|r_i - u| \le 1$:
\begin{equation*}
\text{Var}(\hat{\nabla} J^{DARS-HW}_q) = \text{Var} \left( \sum_{i=1}^{N_{DARS}(u)} \nabla J_i(u) \right) = \sum_{i=1}^{N_{DARS}(u)} \text{Var}(\nabla J_i(u)) = N_{DARS}(u) \Sigma_g = \frac{C}{2u} \Sigma_g.
\end{equation*}
Taking the ratio of the two variances yields:
\begin{equation*}
\frac{\text{Var}(\hat{\nabla} J^{MaxRL}_q)}{\text{Var}(\hat{\nabla} J^{DARS-HW}_q)} = \frac{N/u^2}{C/(2u)} = \frac{2N}{C} \cdot \frac{1}{u} \propto \frac{1}{u}.
\end{equation*}

Therefore, as $u \to 0$, the variance of MaxRL grows at a rate of $\mathcal{O}(1/u^2)$, while DARS-HW grows at a significantly slower rate of $\mathcal{O}(1/u)$.

\subsection{Theoretical Analysis Conclusions}
\label{app:theory_conclusion}
\begin{itemize}
    \item \textbf{Unified Optimization Objective with Divergent Implementation Routes.}
DARS-HW and MaxRL are two distinct implementation routes toward the same Maximum Likelihood reinforcement learning objective. MaxRL achieves the target optimization weight by \textit{algebraically modifying the advantage function} ($\hat{A}^{MaxRL}_i = \frac{r_i - u}{u}$), scaling the gradient signal of each rollout by $1/u$. In contrast, DARS-HW retains the original unscaled advantage function ($\hat{A}_i = r_i - u$) and instead achieves the identical expected gradient scaling by \textit{adaptively increasing the number of rollouts} ($N_{DARS}(u) \propto 1/u$) for difficult problems.
    \item \textbf{Variance Reduction of DARS.}
Appendix \ref{apd:variance} fundamentally explains the superiority of DARS over pure advantage-scaling methods as MaxRL. MaxRL heavily amplifies the gradient of rare, potentially idiosyncratic correct trajectories. This leads to a high-variance gradient regime, which may induce optimization instability and premature convergence. By contrast, DARS-HW functions as a variance-reduced estimator. Instead of amplifying a small fixed set of rollouts algebraically, DARS-HW allocates explicit computational budget to difficult problems, thereby expanding the breadth of \textbf{actual} correct reasoning paths discovered. This mechanism preserves a high signal-to-noise ratio in the gradient and maintains robust token-level entropy, theoretically validating DARS-HW as a more stable and efficient paradigm for Maximum Likelihood RL optimization.
\end{itemize}

%%%%%%%%%%%%%%%%%%%%%%%%%%%%%%%%%%%%%%

% \begin{discussion-block}
% ...
% \end{discussion-block}

\newpage
\section{Training and Evaluation Details}
\label{apd:detail}

\begin{AIbox}{Prompt for Solving Complex Reasoning Tasks}
Your task is to solve the given question step by step. You should conduct a systematic, thorough reasoning process before providing the final answer. This involves analyzing, summarizing, exploring, reassessing, and refining your reasoning process through multiple iterations. Each reasoning step should include detailed analysis, brainstorming, verification, and refinement of ideas. You should include the final answer in \textbackslash boxed\{\} for closed-form results like multiple choices or mathematical results.
\end{AIbox}

\paragraph{Parameters and Metrics.} Currently, our experiments are conducted with Qwen2.5-Math series language models~\citep{qwen25math}. We set the temperature to 1.0 for both the training and evaluation procedures. In this paper, we mainly use two metrics, \textit{Pass@1} and \textit{Pass@K}.
To acquire \textit{Pass@K} results, we sample 128 candidate responses for each question during the evaluation process; the calculation of \textit{Pass@1} is derived from \textit{Avg@128}. Both the training and evaluation processes are scored using Math-Verify.
The learning rate is 1e-6 for depth training methods, and 5e-6 for large breadth training. We do not use the reference model and KL loss. For fair comparison, we uniformly set the PPO mini step to 2 for all experiments. By default, the maximum prompt length is 1024, and the maximum response length is 3072 for the Qwen2.5-Math series model.

\paragraph{Implementation Details.}
Following LUFFY~\citep{luffy}, we use the default subset and filter out generations that are longer than 8192 tokens and those that are verified wrong by Math-Verify \footnote{https://github.com/huggingface/Math-Verify}, resulting in 45k question-solution pairs. 
For training Llama-3.1-8B, we use the train split of MATH dataset.
Our training framework is derived from Verl~\citep{verl} pipeline, which is a flexible, high-performance reinforcement-learning framework built for training large language-model agents. With native PyTorch support and efficient distributed training, Verl lets researchers quickly prototype and scale RL algorithms like PPO on GPUs. Following Dr. GRPO \citep{liu2025understanding}, we remove the KL loss and the length normalization in GRPO.
For rollout generation, we use the modern serving stack provided by \textsc{verl} with optimized backends such as vLLM. These backends employ continuous batching and asynchronous scheduling, which reduce padding waste and mitigate the straggler effect when different prompts receive different rollout budgets.
All of our experiments are conducted on H200 GPUs. At present, the LLM of our experiment is the Qwen2.5-Math series.

\paragraph{Training Steps and Checkpoint Steps.}
For non-breadth methods on Qwen2.5-Math-1.5B/7B, we set the checkpoint step as 100. 
For breadth methods on Qwen2.5-Math-1.5B/7B, we set the checkpoint step as 15.
The specific training steps are determined according to the convergence of the model. The number of training steps for non-breadth training is set as 300 for Llama-3.1-8B, 600 for Qwen2.5-Math-1.5B, and 500 for Qwen2.5-Math-7B. The number of training steps for breadth training is set to 70. For breadth training, we set the total training steps as 105 for Qwen2.5-Math-1.5B, and 75 for Qwen2.5-Math-7B.

\newpage
\section{More Experimental Results}
\label{apd:more_exp}

\subsection{Generalization to Other Task Domains}
\label{apd:ood_domain}
To examine whether DARS is overly specialized to mathematical RLVR, we additionally evaluate the trained Qwen2.5-Math checkpoints on two out-of-domain benchmarks: GPQA-Diamond and HumanEval. These tasks differ substantially from the mathematical training domain and test scientific reasoning and code generation ability, respectively.

As shown in Table \ref{tab:ood_domain}, DARS remains effective beyond mathematics. In particular, DARS-HW-Breadth consistently improves over the RLVR baseline (Dr. GRPO) on GPQA-Diamond \cite{rein2024gpqa} and HumanEval \citep{chen2021evaluating} \textit{Avg@64} for both model scales, while also improving HumanEval \textit{Pass@64}. These results suggest that the core mechanism of DARS, allocating more sampling budget to difficult instances, is largely task-agnostic and can transfer to other reasoning domains.
\begin{table}[htbp]
    \centering
    \small
    \renewcommand\arraystretch{1.3}
    \setlength{\tabcolsep}{6pt}
    \caption{Additional out-of-domain evaluation on non-mathematical reasoning benchmarks.}
    \begin{tabular}{l|c|c|c}
       \toprule
       \textbf{Method} & \textbf{GPQA-Diamond} (\textit{Avg@64}) & \textbf{HumanEval} (\textit{Avg@64}) & \textbf{HumanEval} (\textit{Pass@64}) \\
       \midrule
       \rowcolor{gray!16} \multicolumn{4}{c}{\textit{Qwen2.5-Math-\highlight{1.5B} as the Base Model}} \\
        RLVR Baseline & 28.5 & 41.1 & 84.2 \\
        \rowcolor{table-blue!22}DARS-HW & 28.4 & 40.7 & \textbf{85.8} \\
        \rowcolor{table-blue!66}\textbf{DARS-HW-Breadth} & \textbf{29.3} & \textbf{49.5} & 85.6 \\
        \midrule
        \rowcolor{gray!16} \multicolumn{4}{c}{\textit{Qwen2.5-Math-\highlight{7B} as the Base Model}} \\
        RLVR Baseline & 38.5 & 63.2 & 93.9 \\
        \rowcolor{table-blue!22}DARS-HW & 38.6 & 63.7 & 94.8 \\
        \rowcolor{table-blue!66}\textbf{DARS-HW-Breadth} & \textbf{39.6} & \textbf{68.3} & \textbf{94.9} \\
        \toprule
    \end{tabular}
    \label{tab:ood_domain}
\end{table}

\subsection{Ablation Study on std-based Advantage Computation}
\label{app:std_ablation}
\begin{figure}[htbp] % htbp    % width=0.99\textwidth
    \centering
    \includegraphics[width=0.8\linewidth]{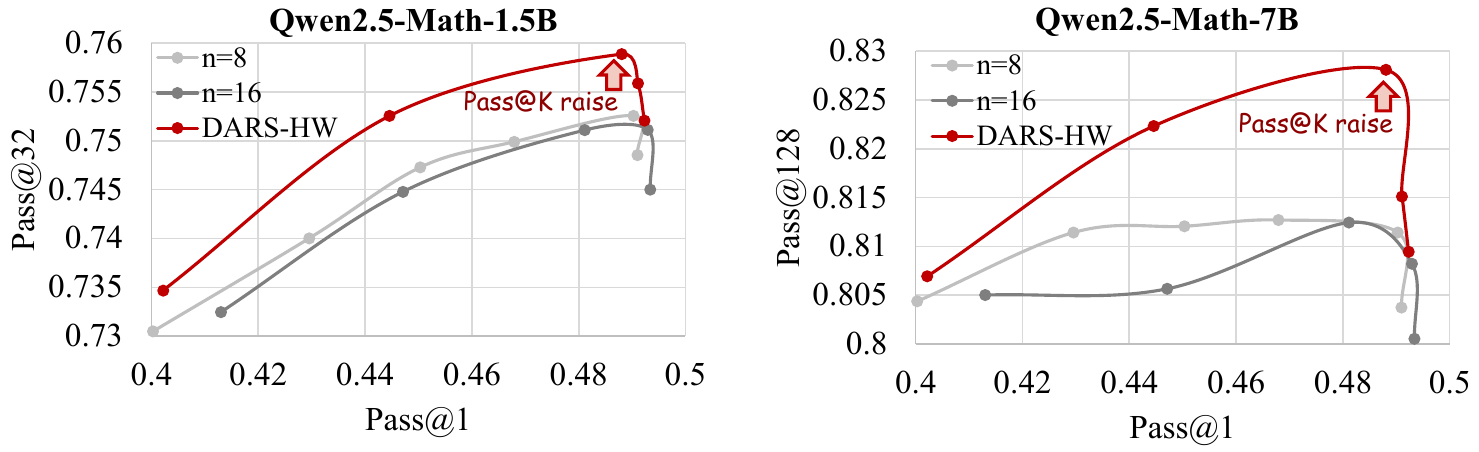}
    % \vspace{-2mm}
    \caption{Comparison of our DARS on std-based advantage computation.}\label{fig:std-dars}
    % \vspace{-6mm}
\end{figure}

As illustrated in Section \ref{sec:analysis}, Dr.~GRPO~ \citep{liu2025understanding} removes the standard-deviation term from the advantage computation to eliminate question-level difficulty bias, and demonstrates its superiority through extensive experiments. Consequently, the experiments reported in this study were conducted primarily though the Dr.~GRPO methodology. To further illustrate the effectiveness of DARS on std-based advantage computation, we conduct the experiment with HW schedule on Qwen2.5-Math-1.5B model, as shown in Figure \ref{fig:std-dars}.
\label{apd:std}

\newpage
\subsection{Comparison with Static Difficulty Allocation}
\label{apd:static_dars}
A natural heuristic baseline is to pre-compute a fixed difficulty score for each training problem using the initial base model, and then keep the rollout budget unchanged throughout RL training. We denote this variant as \textit{Static DARS}. Specifically, we first evaluate the training set with Qwen2.5-Math-1.5B/7B before RL, estimate per-problem difficulty from the base-model accuracy, and assign fixed differentiated rollout budgets with the same HW schedule used by DARS-HW. The results are shown in Table \ref{tab:static_dars}. Static DARS consistently improves over the GRPO baseline and the naive uniform depth scaling baseline, showing that difficulty-aware budget allocation is useful. However, it still underperforms online DARS-HW on \textit{Pass@128} for both model scales. This supports our claim that difficulty is policy-dependent and evolves during RL: once the model masters a subset of questions, a static allocation continues to spend rollout budget on them, whereas online DARS keeps refocusing compute on the current capability frontier.

\begin{table}[htbp]
    % \vspace{-2mm}
    \centering
    \small
    \renewcommand\arraystretch{1.3}
    \setlength{\tabcolsep}{5pt}
    \caption{Comparison with heuristic baselines and a static difficulty-aware rollout allocation baseline. Static DARS computes difficulty scores once using the initial base model and keeps rollout budgets fixed during RL training.}
    \begin{tabular}{l|c|c|c|c|c|c|c}
       \toprule
       \textbf{Method} & \textbf{AIME24} & \textbf{Math500} & \textbf{Olympiad} & \textbf{AMC} & \textbf{Minerva} & \textbf{Avg@128} & \textbf{Pass@128} \\
       \midrule
       \rowcolor{gray!16} \multicolumn{8}{c}{\textit{Qwen2.5-Math-\highlight{1.5B} as the Base Model}} \\
        RLVR Baseline & 14.7 & 75.9 & 39.4 & 47.5 & 31.2 & 49.6 & 79.6 \\
        Depth-Naive (N=32) & 16.5 & 76.2 & 39.9 & 47.9 & 30.9 & 50.1 & 79.9 \\
        \rowcolor{table-blue!12} Static DARS & 17.3 & 76.7 & 40.3 & 47.3 & 31.2 & 50.3 & 80.4 \\
        \rowcolor{table-blue!62}\textbf{DARS-HW} & 17.7 & 76.4 & 40.0 & 48.4 & 30.8 & 50.0 & \textbf{82.1} \\
        \midrule
        \rowcolor{gray!16} \multicolumn{8}{c}{\textit{Qwen2.5-Math-\highlight{7B} as the Base Model}} \\
        RLVR Baseline & 26.8 & 82.2 & 44.3 & 57.2 & 35.7 & 55.3 & 81.4 \\
        Depth-Naive (N=32) & 28.0 & 83.8 & 46.4 & 59.0 & 37.3 & 57.0 & 80.3 \\
        \rowcolor{table-blue!12} Static DARS & 28.6 & 84.0 & 46.7 & 59.7 & 37.6 & 57.4 & 81.1 \\
        \rowcolor{table-blue!62}\textbf{DARS-HW} & 30.1 & 83.5 & 47.1 & 59.4 & 37.2 & 57.3 & \textbf{83.5} \\
        \toprule
    \end{tabular}
    \label{tab:static_dars}
\end{table}

\newpage

\subsection{Training Dynamics Comparison of DARS-ET/HW and Depth-Naive}
\label{app:training_dynamics_compare}
\begin{figure*}[htbp] % htbp    % width=0.99\textwidth
    \centering
    % \vspace{-5mm}
    \includegraphics[width=0.9\linewidth]{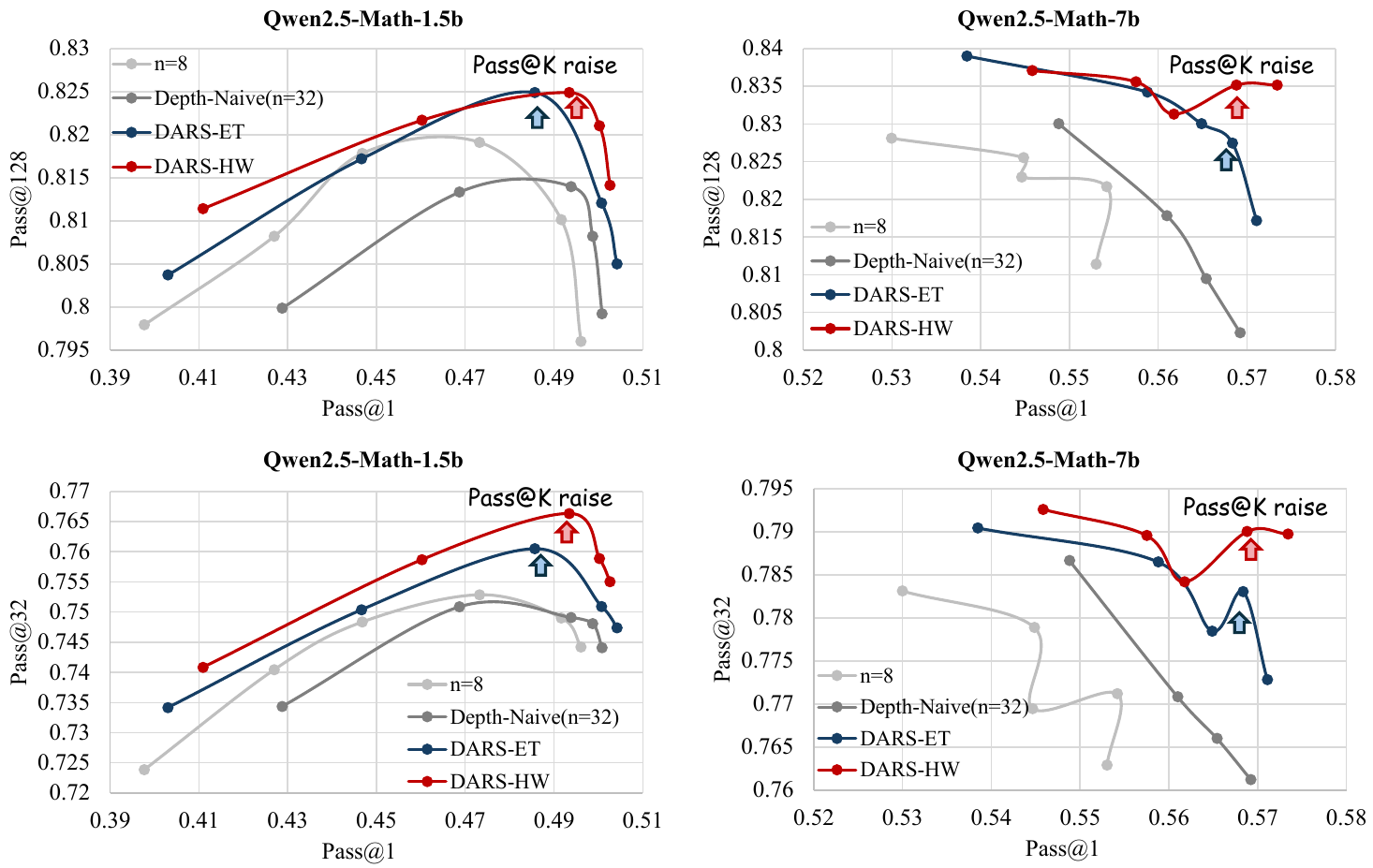}
    % \vspace{-6mm}
    \caption{Training dynamics of \textit{Pass@32}/\textit{Pass@128} and \textit{Pass@1} with different training steps of Qwen2.5-Math-1.5B and -7B.}\label{fig:depth-capacity}
    % \vspace{-3mm}
\end{figure*}

We show this training dynamics comparison for Qwen2.5-Math-Series in Fig \ref{fig:depth-capacity}. The training dynamics illustrate the effectiveness of both DARS-ET and DARS-HW.

\subsection{Depth and Breadth Synergy for Pass@1 and Pass@32}
\label{apd:pass32-1}
\begin{figure}[htbp] % htbp    % width=0.99\textwidth
    \centering
    \vspace{-2mm}
    \includegraphics[width=0.9\linewidth]{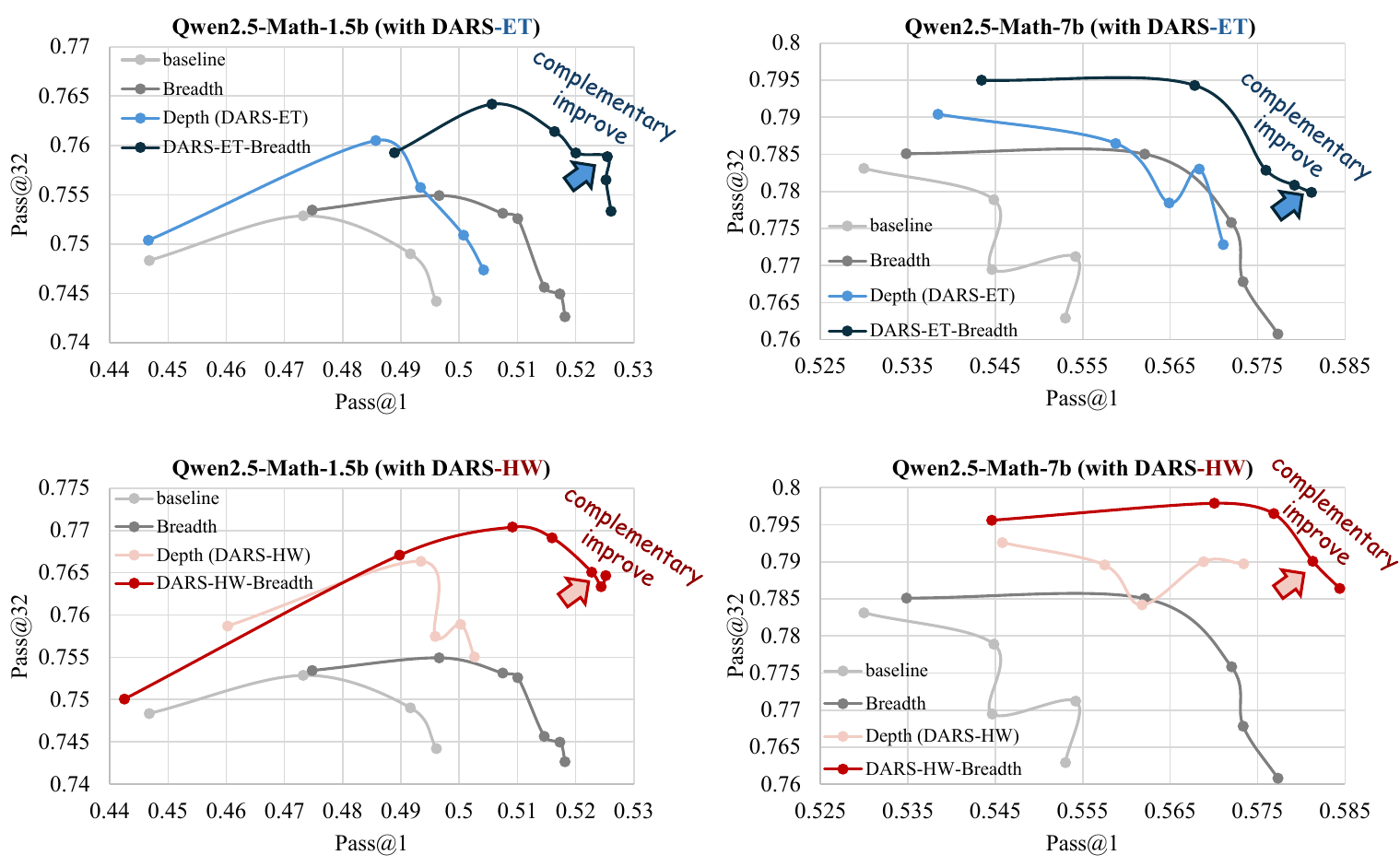}
    \caption{Complementary improve of Depth and Breadth Synergy for \textit{Pass@1} and \textit{Pass@K} (K=32) performance.}
    \label{fig:D_B_32}
    \vspace{-2mm}
\end{figure}

In Section \ref{sec:breadth-method}, we show the training dynamics of \textit{Pass@128}-\textit{Pass@1} for DARS and baseline methods. To further illustrate the effectiveness of DARS, we show the training dynamics of \textit{Pass@32}-\textit{Pass@1} in Figure \ref{fig:D_B_32}. Our DARS significantly improves the \textit{Pass@32} performance compared to other methods.

\begin{figure}[htbp] % htbp    % width=0.99\textwidth
    \centering
    \includegraphics[width=0.9\linewidth]{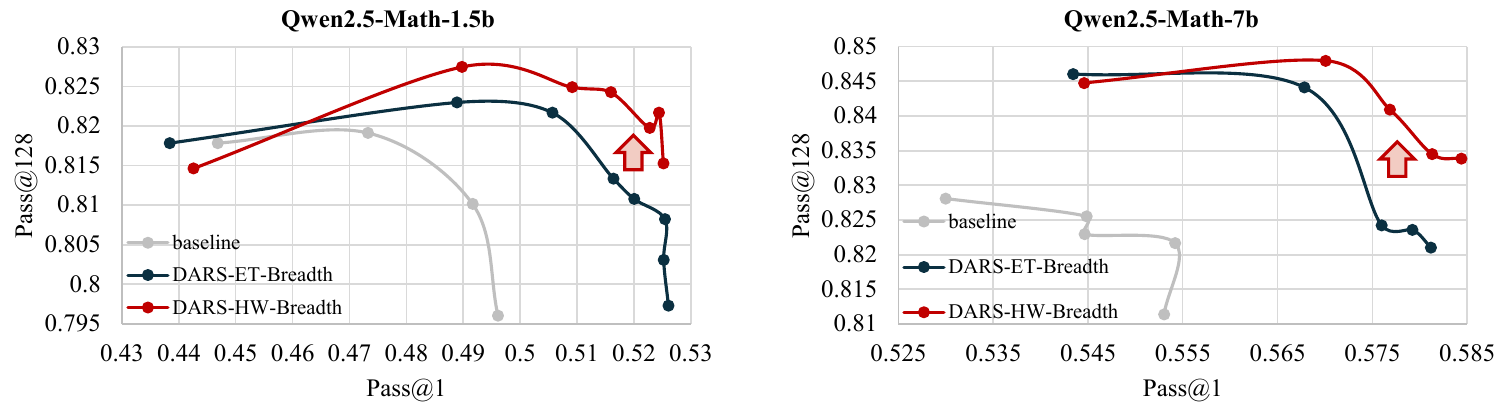}
    \caption{Comparison of ET and HW schedule in breadth training of Qwen2.5-Math series.}\label{fig:et-he}
\end{figure}

\newpage
\subsection{Comparison of ET/HW Schedule in Breadth Scaling}
\label{app:et_hw_breadth}
In addition, compared with the ET schedule, DARS-HW-Breadth significantly improves the model's Pass@128 performance as shown in Figure \ref{fig:et-he}. We consider this performance gain is due to the HW schedule placing greater emphasis on difficult samples.

\subsection{Consistent Improvement During RL Process}
\label{app:consistent_improve_rl}
% The primary goal of this paper is to ensure no regression in Pass@1 while maximizing Pass@128.
To further show that our method consistently improve model performance, we calculated the mean of Pass@128 and Pass@32 for the last 3 checkpoints of each method, as shown in Table \ref{tab:ckpt}.
\vspace{-2mm}
\begin{table}[htbp]
    \centering
    \small
    % \scriptsize
    \renewcommand\arraystretch{1.3}
    \setlength{\tabcolsep}{6pt}
    \caption{Average performance of \textit{Pass@1/32/128} for the last 3 checkpoints during training.}
    \begin{tabular}{l|c|c|c}
       \toprule
       \textbf{Model} & Pass@1 & Pass@32 & Pass@128 \\
       \midrule
       \rowcolor{gray!16} \multicolumn{4}{c}{\textit{Qwen2.5-Math-\highlight{1.5B} as the Base Model}} \\
        RLVR Baseline & 48.8 & 74.7 & 80.8 \\
        Depth-Naive & 49.5 & 74.4 & 80.6 \\
        Breadth-Naive & 51.4 & 74.4 & 79.8 \\
        \rowcolor{table-blue!22}DARS-HW & 49.5 & 75.7 & 81.9 \\
        \rowcolor{table-blue!66}\textbf{DARS-HW-Breadth}& \textbf{52.4} & \textbf{76.4} & \textbf{82.1} \\
        \midrule
        \rowcolor{gray!16} \multicolumn{4}{c}{\textit{Qwen2.5-Math-\highlight{7B} as the Base Model}} \\
        RLVR Baseline & 55.1 & 76.9 & 81.8 \\
        Depth-Naive & 56.4 & 76.7 & 80.9 \\
        Breadth-Naive & 57.2 & 76.7 & 81.3 \\
        \rowcolor{table-blue!22}DARS-HW & 56.8 & 78.8 & 83.4 \\
        \rowcolor{table-blue!66}\textbf{DARS-HW-Breadth} & \textbf{58.3} & \textbf{79.1} & \textbf{83.7} \\
        \toprule
    \end{tabular}
    \label{tab:ckpt}
\end{table}

\newpage
\subsection{Performance of Other Model}
\label{app:other_model}
We further evaluate our method on Qwen2.5-7B-Instruct. Following \citep{liang2025beyond}, we change the training data to Math12k. The results are shown in Table \ref{tab:none_math}. 

We further evaluate DARS with the openPangu-7B architecture on Huawei Ascend NPU, to investigate the generalization ability of DARS on different models and hardware architectures. The experimental results are reported in Table \ref{tab:pangu}, and DARS improves the performance of the openPangu model consistently on all datasets. 

As the results show, our method still outperforms the baseline in both the Pass@1 and Pass@K metrics.

\begin{table}[htbp]
    \centering
    \small
    % \scriptsize
    \renewcommand\arraystretch{1.3}
    \setlength{\tabcolsep}{4pt}
    \caption{Overall performance of \textit{Pass@1} (\textit{Avg@128}) and \textit{Pass@128} of Qwen2.5-7B-Instruct.}
    \begin{tabular}{|l|c|c|c|c|c|c|c|}
       \toprule
       \textbf{Model} & AIME24 & Math500 & Olympiad & AMC & Minerva & Avg@128 & Pass@128 \\
       \midrule
        Qwen2.5-7B-Instruct & 11.9 & 72.3 & 37.1 & 42.2 & 31.9 & 47.2 & 80.3 \\
        RLVR baseline & 14.2 & 74.8 & 37.6 & 43.4 & 33.4 & 48.6 & 78.8 \\
        \rowcolor{table-blue!66}\textbf{DARS-HW-Breadth} & 15.6 & 76.5 & 38.4 & 44.7 & 34.6 & 49.6 & \textbf{82.3} \\
        % \midrule
        % openPangu-7B & 36.99 & 82.71 & 48.82 & 51.61 & 37.0 & 57.54 & 79.92 \\
        % \rowcolor{table-blue!66}\textbf{openPangu-7B + DARS} & \textbf{60.58} & \textbf{93.95} & \textbf{67.96} & \textbf{81.06} & \textbf{43.4} & \textbf{72.56} & \textbf{86.59} \\
        \toprule
    \end{tabular}
    \label{tab:none_math}
\end{table}
\begin{table}[htbp]
    \centering
    \small
    % \scriptsize
    \renewcommand\arraystretch{1.3}
    \setlength{\tabcolsep}{4pt}
    \caption{Overall performance of DARS with openPangu on Ascend NPU.}
    \begin{tabular}{|l|c|c|c|c|c|c|c|}
       \toprule
       \textbf{Model} & AIME24 & Math500 & Olympiad & AMC & Minerva & Avg@32 & Pass@32 \\
       \midrule
        openPangu-7B & 36.99 & 82.71 & 48.82 & 51.61 & 37.0 & 57.54 & 79.92 \\
        \rowcolor{table-blue!66}\textbf{openPangu-7B + DARS} & \textbf{60.58} & \textbf{93.95} & \textbf{67.96} & \textbf{81.06} & \textbf{43.4} & \textbf{72.56} & \textbf{86.59} \\
        \toprule
    \end{tabular}
    \label{tab:pangu}
\end{table}

\subsection{Explanation of Significant Pass@k Gain for Llama-3}
\label{app:llama3_passk}
The stronger gains for Llama-3 likely stem from its base model characteristics. Recent studies \citep{gandhi2025cognitive,wang2025octothinker} suggest that base models like Llama-3 lack mid-training, causing many valid responses to appear only at higher sampling budgets. As shown in the table below, Llama-3-8B-Base exhibits a much larger performance gap when increasing (K) from 8 to 32/64 compared to Qwen models as shown in Table \ref{tab:explanation}. This indicates that Llama-3 has more correct solutions “hidden” deeper in the sampling space. Our DARS method, by focusing on hard examples, effectively uncovers these “deep” correct samples, leading to more substantial Pass@K improvements for Llama-3.

\begin{table}[htbp]
    \centering
    \small
    % \scriptsize
    \renewcommand\arraystretch{1.3}
    \setlength{\tabcolsep}{4pt}
    \caption{Pass@K gap comparison of Qwen2.5-Math-1.5/7B and LLama3-8B-Base}
    \begin{tabular}{|l|c|c|c|}
       \toprule
       \textbf{Model} & Pass@8 & Pass@32 - Pass@8 & Pass@64 - Pass@8 \\
       \midrule
        Qwen2.5-Math-1.5B & 56.5 & 12.4 & 17.1 \\
        Qwen2.5-Math-7B & 63.5 & 10.5 & 14.4 \\
        \rowcolor{table-blue!66}\textbf{Llama3-8B-Base} & 17.2 & \textbf{19.1} & \textbf{27.3} \\
        \toprule
    \end{tabular}
    \label{tab:explanation}
\end{table}

\newpage
\subsection{Thinking Length Dynamics}
\label{app:thinking_length}
\begin{figure}[htbp] % htbp    % width=0.99\textwidth
    \centering
    \includegraphics[width=0.9\linewidth]{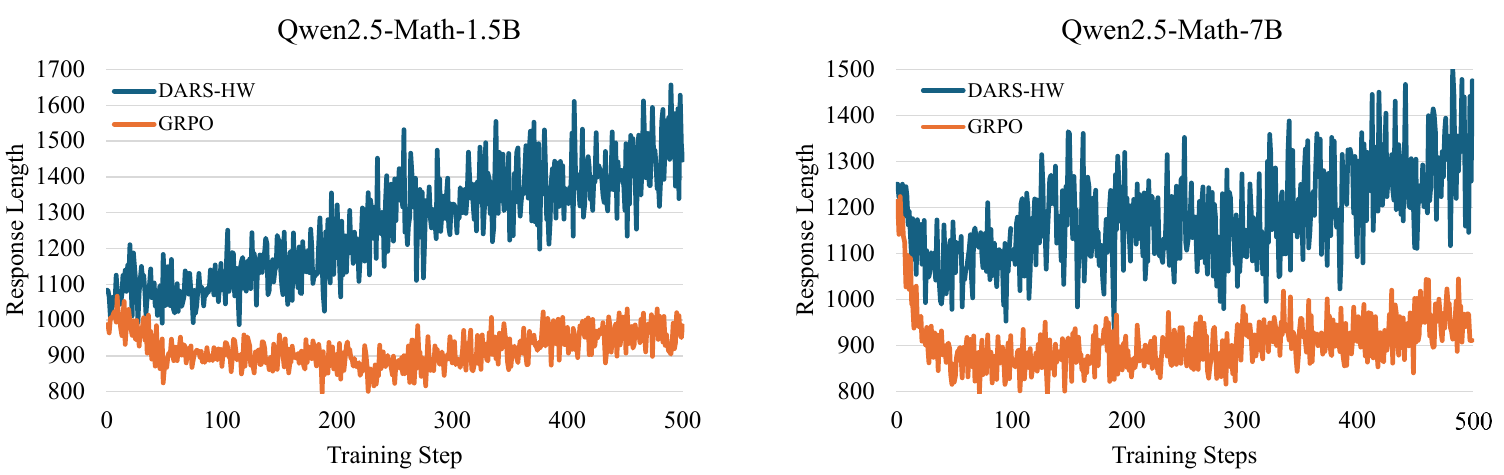}
    % \vspace{-2mm}
    \caption{Training dynamics of response length for GRPO and DARS.}
    \label{fig:train_length}
    % \vspace{-4mm}
\end{figure}

\begin{figure}[htbp] % htbp    % width=0.99\textwidth
    \centering
    \includegraphics[width=0.9\linewidth]{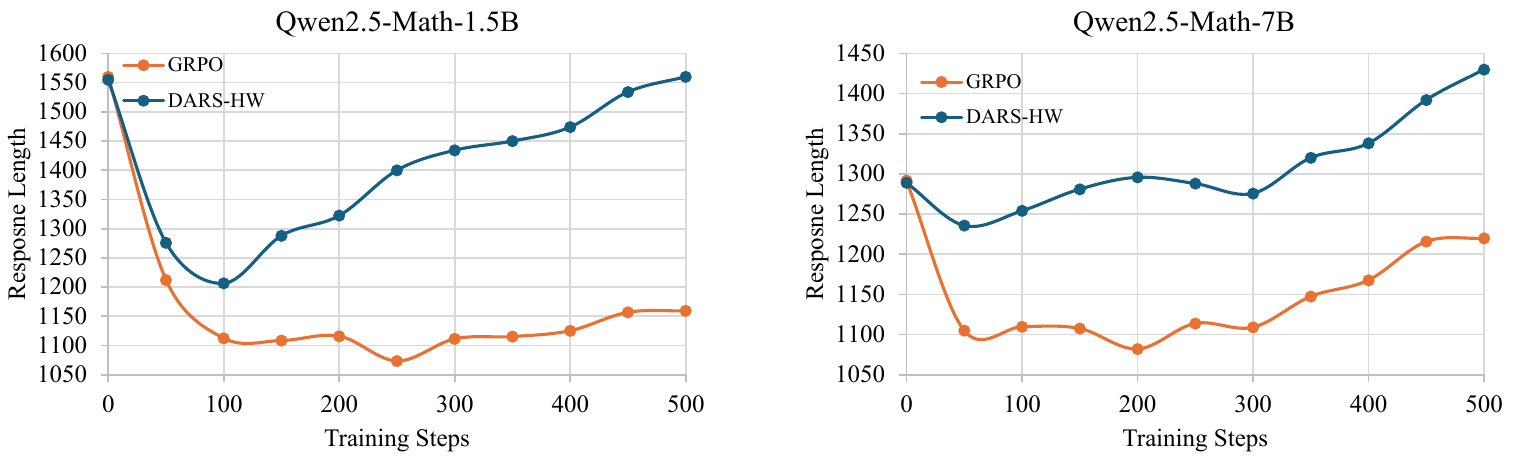}
    % \vspace{-2mm}
    \caption{Statistical results of response length on AIME 2024 for GRPO and DARS.}
    \label{fig:aime_length}
    % \vspace{-4mm}
\end{figure}

This section investigates how DARS influences the reasoning length of LLMs. We tracked the response length dynamics during the training of Qwen2.5-Math-1.5B and 7B models. Our experiments reveal two key observations: (1) The training process shows a clear trend of increasing generation length, as shown in Figure \ref{fig:train_length}. (2) When evaluated on AIME 2024, models trained with DARS consistently produce longer reasoning traces than the baseline, as shown in Figure \ref{fig:aime_length}. (For steps in multiples of 100, we take the average of 128 sampling results. To increase the density of data points, we also sample 16 times for every intermediate 50-step interval and take the average.) 
These results provide concrete evidence that our DARS method successfully stimulates the model to perform deeper and more thorough thinking.

\newpage
\newpage
\section{Case Study}
\label{app:case_study}
\begin{figure}[htbp] % htbp    % width=0.99\textwidth
    \centering
    \includegraphics[width=0.99\linewidth]{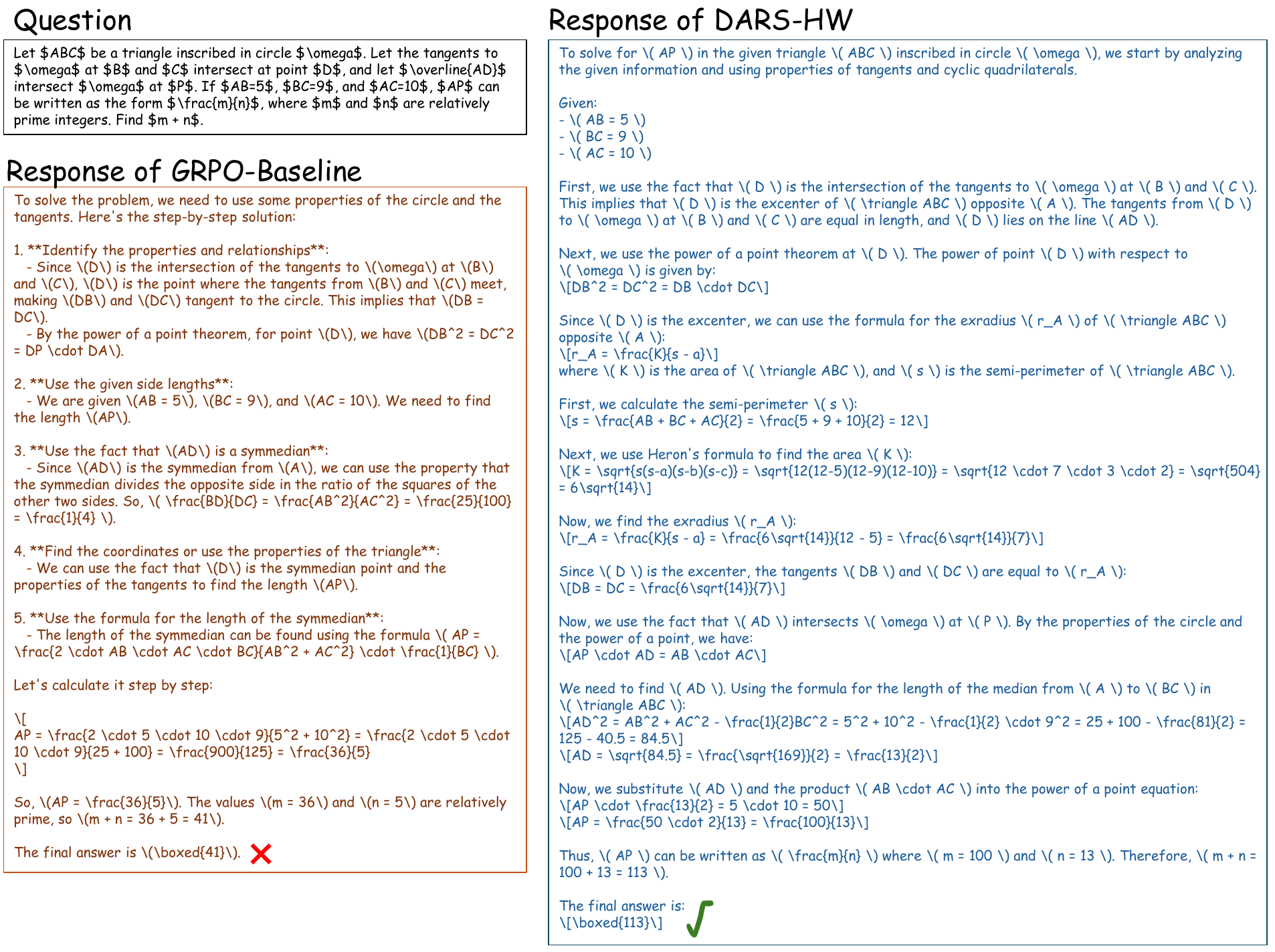}
    % \vspace{-2mm}
    \caption{A case study to demonstrate how DARS solves hard problems that the GRPO baseline misses. This is example is extracted from AIME24 for Qwen2.5-Math-7B.}
    \label{fig:case}
\end{figure}

We show an example of AIME24 to compare the difference for GRPO/DARS-HW trained model. As shown in Fig \ref{fig:case}, compared to GRPO trained model, DARS-HW trained model will generate longer response to solve the hard question.

\newpage
\section{Discussion and Future Work}
\label{app:discussion_future}
In this section, we analyze how hyperparameters \(N\) and \(N^{\max}\) control the shape of the cumulative advantage curve, and how this shape may influence training behavior. We further discuss how dynamically adjusting these parameters could enable a smooth transition from \textit{Pass@K}-oriented to \textit{Pass@1}-oriented training.

\subsection{Hyperparameter Control of Cumulative Advantage Shape}
\label{apd:hyperparam}
\begin{figure}[htbp] % htbp    % width=0.99\textwidth
    \centering
    \includegraphics[width=0.9\linewidth]{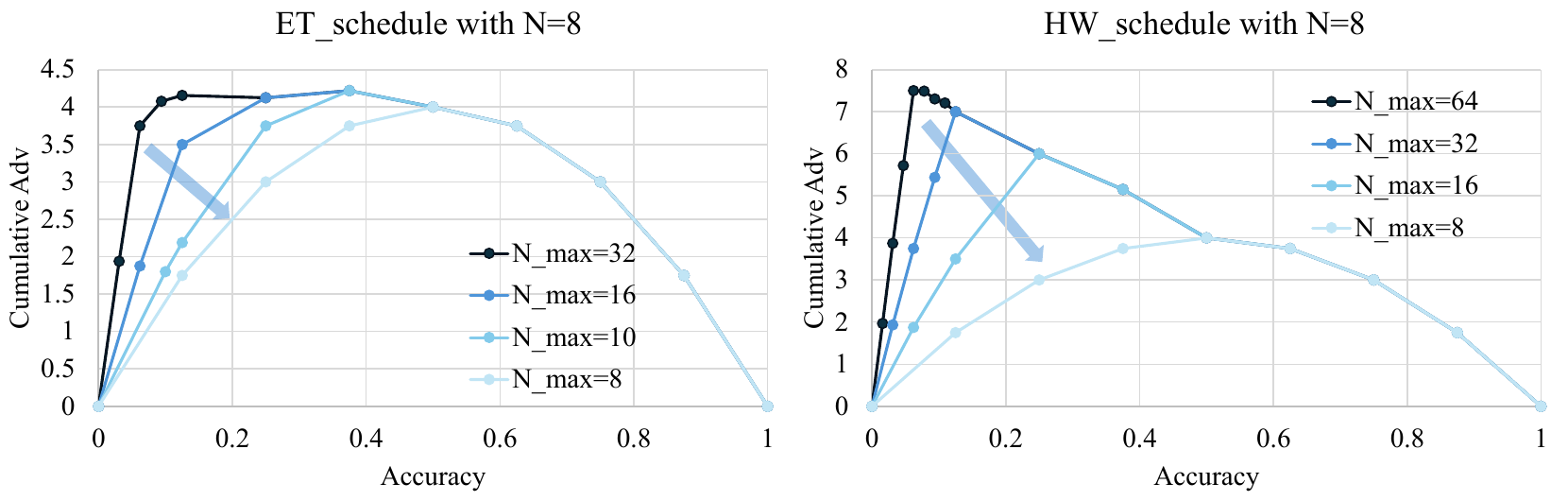}
    % \vspace{-2mm}
    \caption{Control the shape of Cumulative Advantage by adjusting the $N_{max}$ hyperparameter of DARS.}
    \label{fig:cum_adv_shape}
\end{figure}

We show the Cumulative Advantage shape of ET/HW schedule with $N=8$ in Figure \ref{fig:cum_adv_shape}. By continuously reducing the size of $N_{\text{max}}$, the curve will contract accordingly. When $N_{\text{max}} = N$, it is equivalent to the vanilla method without DARS.

\subsection{Potential Pass@K to Pass@1 Training Transition}
\label{app:passk_to_pass1_transition}
The dynamic control of \(N^{\max}\) suggests an intriguing training strategy: starting with a large \(N^{\max}\) value to maximize \textit{Pass@K} performance through extensive exploration of hard problems, then gradually reducing \(N^{\max}\) throughout training to transition toward \textit{Pass@1} optimization.
This approach mirrors curriculum learning principles, where the training difficulty is progressively adjusted. Initially, the model benefits from the expanded solution space and diverse reasoning patterns discovered through heavy sampling on hard problems (high \(N^{\max}\)). As training progresses and the model's capability matures, reducing \(N^{\max}\) focuses the training on refining the most promising solution strategies, ultimately improving single-shot performance.
% Future work will explore optimal annealing schedules for \(N^{\max}\) and investigate whether this transition strategy can simultaneously maximize both \textit{Pass@1} and \textit{Pass@K} performance, potentially overcoming the current limitations of RLVR training.

To preliminarily validate this hypothesis, we conducted a new experiment on Qwen2.5-Math-1.5B with DARS-HW using an \(N^{\max}\) annealing schedule. Specifically, we kept a high rollout cap in the early stage to emphasize exploration on hard problems, and then reduced the cap later in training to encourage convergence to the best solution paths. Concretely, we set \(N^{\max}=64\) for the first 200 training steps, and then linearly annealed it to \(16\) during the remaining training steps.

Table \ref{tab:nmax_anneal} shows that, compared with the fixed-budget DARS-HW setting (\(N^{\max}=64\) throughout training), the annealed variant slightly reduces \textit{Pass@128} from \(82.1\) to \(80.6\), but improves \textit{Pass@1} (\textit{Avg@128}) from \(50.0\) to \(50.9\). This preliminary result is consistent with our hypothesis: a large early-stage \(N^{\max}\) helps discover diverse valid trajectories, while a smaller late-stage \(N^{\max}\) helps the policy concentrate on its best reasoning paths. Future work will explore more principled annealing schedules for \(N^{\max}\) and investigate whether this transition strategy can simultaneously maximize both \textit{Pass@1} and \textit{Pass@K} performance, potentially overcoming the current limitations of RLVR training.
% \vspace{-4mm}

\begin{table}[htbp]
    \centering
    \small
    \renewcommand\arraystretch{1.3}
    \setlength{\tabcolsep}{6pt}
    \caption{Preliminary study of an \(N^{\max}\) annealing schedule for transitioning from \textit{Pass@K}-oriented exploration to \textit{Pass@1}-oriented refinement.}
    \begin{tabular}{l|c|c|c|c}
       \toprule
       \textbf{Method} & \textbf{Initial \(N^{\max}\)} & \textbf{Final \(N^{\max}\)} & \textbf{Pass@128} & \textbf{Pass@1 (Avg@128)} \\
       \midrule
       DARS-HW (Fixed) & 64 & 64 & \textbf{82.1} & 50.0 \\
       \rowcolor{table-blue!22} DARS-HW (Annealed) & 64 & 16 & 80.6 (\(\downarrow\) 1.5) & \textbf{50.9} (\(\uparrow\) 0.9) \\
       \toprule
    \end{tabular}
    \label{tab:nmax_anneal}
\end{table}

% \vspace{-4mm}

\end{document}